\RequirePackage{fix-cm}
\documentclass[twocolumn]{svjour3}           
\smartqed  
\usepackage{graphicx}
\usepackage{color}
\usepackage{url}
\usepackage{hyperref}

\usepackage{color}
 \usepackage{multirow}
 \usepackage[cmex10]{amsmath}
 
 \usepackage[colorinlistoftodos]{todonotes} 

\begin{document}

\title{Real-Time Visual Place Recognition for Personal Localization on a Mobile Device
\thanks{This research was funded by the National Science Centre in Poland in years 2016-2019 under
the grant 2015/17/N/ST6/01228. M. Nowicki received doctoral scholarship from the National
Science Centre in Poland under the grant 2016/20/T/ST7/00396}}


\author{Micha{\l} Nowicki \and Jan Wietrzykowski \and Piotr Skrzypczy\'nski}


\institute{The authors are with the Institute of Control and Information Engineering,
           Pozna\'n University of Technology \\
           ul. Piotrowo 3A, PL-60-965 Poza\'n, Poland \\
           Tel.: +48-61-6652365\\
           Fax: ++48-61-6652563\\
           \email{name.surname@put.poznan.pl}     
}

\date{Received: date / Accepted: date}

\maketitle

\begin{abstract}
The paper presents an approach to indoor personal localization
on a mobile device based on visual place recognition. We implemented on a
smartphone two state-of-the-art algorithms that are representative
to two different approaches to visual place recognition: FAB-MAP that
recognizes places using individual images, and ABLE-M that utilizes
sequences of images. These algorithms are evaluated in environments of
different structure, focusing on problems commonly encountered when a mobile
device camera is used. The conclusions drawn from this evaluation are guidelines
to design the FastABLE system, which is based on the ABLE-M algorithm,
but introduces major modifications to the concept of image matching.
The improvements radically cut down the processing time and improve scalability,
making it possible to localize the user in long image sequences with the
limited computing power of a mobile device. The resulting place recognition
system compares favorably to both the ABLE-M and the FAB-MAP solutions in
the context of real-time personal localization.
\keywords{Visual place recognition \and indoor localization \and mobile localization}
\end{abstract}

\vspace{-3mm}
\section{Introduction}
\vspace{-2mm}
\subsection{Motivation}
Personal indoor localization is still a problem to be solved without using an
additional infrastructure. A~number of emerging applications of mobile devices
require to localize the user with respect to predefined objects or places.
Although much work has been done on practical localization in GPS-denied indoor
environments, resulting in commercially available 
solutions\footnote{\url{http://www.skyhookwireless.com/}}, there is a need for 
approaches that are independent of the artificial sources of signals, such as 
RFID/BLE beacons or WiFi networks. Whereas external infrastructure can be easily 
applied for localization in limited areas~\cite{eaai01}, many applications 
benefit from infrastructure-free indoor localization. Examples range from 
guidance of visitors in large buildings or exhibition areas to location-aware 
advertising in shopping centres \cite{recom}. Nowadays, smartphones and tablets 
have the considerable computing power and are equipped with high-quality cameras. The ever increasing performance of such mobile devices allows us to consider 
vision-based algorithms for personal indoor localization. Localization utilizing 
visual information can operate regardless of the availability of particular 
infrastructure or the presence of specific features in the environment. In 
particular, visual localization by place recognition is similar to the way people and animals localize themselves in a surrounding environment, associating the 
current visual perception to the past episodic memories \cite{redish}.

\subsection{State of the Art}
The application of cameras for localization was already researched extensively
in robotics, resulting in a number of Visual Odometry (VO)~\cite{vo}
and Simultaneous Localization and Mapping (SLAM) \cite{vision} algorithms.
Unfortunately, the use of VO/SLAM on mobile devices is challenging due to the rapid,
unconstrained motion of the embedded camera, limited availability of point features,
and the high computing power demanded by VO/SLAM systems. Among the state-of-the-art
visual localization algorithms, the Parallel Tracking and Mapping (PTAM) algorithm \cite{ptam}
was implemented on iPhone \cite{iptam}, whereas the SVO system~\cite{SVO} was demonstrated
working in real-time on the limited-power Odroid platform. However, the iPhone
implementation of PTAM has limited functionality with only the camera pose tracking
running in real-time in small workspaces, which made this system more suitable for
augmented reality applications than for personal localization in buildings. Also the
SVO, which was developed for small aerial vehicles, is rather unsuitable for personal localization,
as it requires a very fast camera and makes strong assumptions as to the camera motion~\cite{SVO}.

More typical monocular SLAM algorithms, such as the recent ORB-SLAM \cite{orbslam} that
demonstrated very accurate camera trajectory estimation on several benchmarks, cannot be
used in real-time on mobile devices due to the high computing power requirements of their
underlying graph optimization techniques. Although the SlamDunk system, also based on
graph optimization has been re-implemented on a mobile device \cite{dunk},
it utilizes RGB-D data to simplify 3-D perception and requires a Kinect-type
sensor connected to the device, which is infeasible for most applications.
Whereas new sensors aimed at enhancing mobile devices with RGB-D perception,
such as the Intel RealSense become available, this technology still offers
relatively short-range perception. So far it is suitable rather for 3-D object
recognition or augmented reality than for building-scale localization.

Our experience from implementing a typical monocular VO pipeline on
Android-based devices \cite{iciar} suggests that robust, real-time operation of purely
visual odometry is possible only if the acquired images do not show significant motion blur,
and the device held in hand makes no sudden orientation changes. This, together with the
high computation burden of the motion estimation algorithms \cite{iciar} makes VO
impractical for personal indoor localization. Although the recently developed VO algorithms
based on visual and inertial information~\cite{RobustVO,VOKalman} can overcome to a
great extent the problems related to sudden motions of a handheld camera, they still
need relatively high computing power \cite{MScVO}, and thus drain a lot of energy
from the device's battery. Vision-based tracking was already integrated with WiFi
and inertial data for personal localization \cite{thrun}, but using a setup that
only resembled a mobile device, thus avoiding the computing power and energy efficiency issues.

Another possibility in visual localization is the ap\-pear\-ance\--based place recognition~\cite{placsurv}.
Contrary to the VO/SLAM algorithms, the place recognition methods only determine if the observed
place is similar to an already visited location.
However, the place recognition methods scale better for large environments than typical
SLAM algorithms~\cite{loop}. In the ap\-pear\-ance\--based methods each image is processed
and described by descriptors of salient features, or directly described by a vector of
numerical values constituting a whole-image descriptor~\cite{gist}. Direct matching of
local image features is inefficient for place recognition \cite{placsurv}, thus the
Bag of Visual Words (BOVW) technique \cite{bag} is commonly used, which organizes the
features into a visual vocabulary. Images described by visual words can be efficiently
matched using a comparison of binary string or histograms. One prominent example of a
place recognition algorithm that employs the BOVW technique is FAB-MAP~\cite{fabmap1},
which implements efficient comparison of images with a histogram-based method. This
algorithm and its further developments have demonstrated impressive results over very
long sequences of images in various environments \cite{fabmap2}.

Recent approaches to direct image description involve usage of image sequences~\cite{seqslam}.
The use of sequences instead of individual images exploits the temporal coherence of vision
data acquired by a moving camera \cite{sigindoor}, thus decreasing the number of false positives
in place recognition for environments with self-similarities, and increasing the tolerance
to local scene changes. This idea is used by the ABLE-M~\cite{able} algorithm, which
substantially scales down the processed images, and compares global binary descriptors
using the quick to compute Hamming distance.

Image retrieval techniques similar to those applied in ap\-pear\-ance\--based localization
were already used on smartphones in the context of object recognition \cite{wang}.
However, only a few papers tackle the problem of ap\-pear\-ance\--based localization on mobile devices.
An algorithm that combined WiFi fingerprinting without a pre-surveyed database of fingerprints,
and a simple visual ap\-pear\-ance\--based location verification technique was presented in \cite{jamris}.
This work demonstrated that although the ap\-pear\-ance\--based recognition on a smartphone returned a
higher number of false positives than the WiFi-based method, the number of locations correctly
identified using the visual data was still much bigger than the number of locations identified
by matching WiFi fingerprints. If a database of images acquired at known positions is available,
it is possible to compute user position and orientation using an image acquired with the smartphone
camera, as demonstrated in \cite{feng}. However, this approach seems to have a limited potential in
practical applications, as collecting the database of images requires a special trolley with a
multi-sensor set up for accurate localization. To the best knowledge of the authors, their
recent paper~\cite{IPIN} was the first one systematically investigating vision-only ap\-pear\-ance\--based
localization on an actual smartphone, and demonstrating the feasibility of an algorithm based
on sequences of images for personal localization.

\subsection{Contribution}
In this article we significantly extend our preliminary research, presented in
the recent conference paper \cite{IPIN}. This research revealed that the ABLE-M
algorithm based on image sequences is particularly beneficial for place 
recognition in corridor-like environments, where not enough salient features are 
available for reliable comparison based on a single query image. Moreover, the 
local binary descriptors used in ABLE-M make it very fast at computing individual matches between images.
This makes ABLE-M an interesting choice for implementation on mobile devices
for the use in large buildings. However, as we show using the experimental data,
and confirm by the theoretical analysis, the processing time of the ABLE-M 
algorithm does not scale well for long database sequences. Therefore, we propose a much faster
version of the ABLE-M algorithm, called FastABLE, that suits the needs of 
personal localization on mobile devices. In the improved version, the processing 
time is independent of the size of the comparison window. The FastABLE code is 
publicly available as open source, together with the test sequences, to make our 
results fully verifiable.
Thus, the main contribution of this work is twofold:
(i) thorough experimental and theoretical analysis of the properties of the two
representative ap\-pear\-ance\--based place recognition algorithms in the context of implementation on a mobile device;
(ii) new ap\-pear\-ance\--based localization algorithm tailored specifically to 
the requirements of the mobile devices with limited computing power and typical
characteristics of indoor environments.

\section{FAB-MAP}
\label{fabmapa}
The FAB-MAP is a system for visual place recognition proposed by Cummins and Newman~\cite{fabmap1}.
The FAB-MAP algorithm is available in the open-source implementation named OpenFABMAP\footnote{\url{https://github.com/arrenglover/openfabmap}}.
Proposed improvements for FAB-MAP~\cite{fabmapconc,fabmap2} in terms of scalability and evaluation
on large datasets made the system a popular choice for loop closing in visual SLAM systems.

To determine whether the current image represents an already visited or an a priori known place,
FAB-MAP detects and describes point features found in every image.
The computed descriptors are then used to compare images resulting in a system more robust than direct
pixel comparisons that are susceptible to local lighting changes or changes in the observation angle.
For each image, the Bag of Visual Words approach is used to create a histogram that describes the
occurrence of each descriptor in the image.
Even though the descriptors could be directly compared, FAB-MAP uses a probabilistic model that describes
the probability $P(L_i | \mathcal{Z}^k)$ of re-observing the place $L_i$ based on histograms of all images in the database
$\mathcal{Z}$ and a histogram of a current (numbered as $k$-th) frame $Z_k$. The set of those histograms will be further
denoted by $\mathcal{Z}^k = \mathcal{Z} \cup Z_k$.
Comparing histograms using probabilistic model allows distinguishing the features that are relevant in
place recognition and features that are commonly found in the environment.
In the FAB-MAP the probabilistic distribution of features occurrences is approximated by a Chow-Liu tree \cite{chowliu},
allowing efficient computation of the $P(L_i | \mathcal{Z}^k)$ values.

The FAB-MAP algorithm was extensively tested in the outdoor environment, especially with omnidirectional cameras,
and proved to be a very useful component of robotic SLAM systems.
Usually, outdoor datasets contain thousands of features allowing to learn large BOVW vocabulary.
As wheeled mobile robots usually move in a stable and smooth manner, the FAB-MAB was usually run on non-blurred images.
Because of the presented issues, it is unclear if the FAB-MAP algorithm will work well in the case of a shorter,
indoor sequences with motion blur. Moreover, the OpenFABMAP implementation operates only on floating-point descriptors
like SIFT or SURF and thus cannot benefit from the recently developed binary descriptors, which are much faster to
compute and compare.

\subsection{Training and recognition}
Before the FAB-MAP is ready to recognize places, pre-training and training process must be performed.
In the pre-training process, the system determines a types of descriptors that can be found in the environment by creating
the BOVW vocabulary. Therefore, the pre-training images should represent the characteristic of the target operational environment.
In this step, all descriptors found in the pre-training images are clustered using the k-means algorithm with the
Euclidean distance measure. Each cluster constitutes a single word from the center of belonging members which is then inserted in the vocabulary.
This vocabulary is used in a Chow-Liu training procedure that is performed with Meil\v{a}'s algorithm \cite{chowliulearn}.
In turn, in the training process the taken images are processed to create a database of places for recognition.
For each image, the BOVW is computed that is used in the recognition phase.

\begin{figure}[thpb!]
\centering
 \includegraphics[width=\columnwidth]{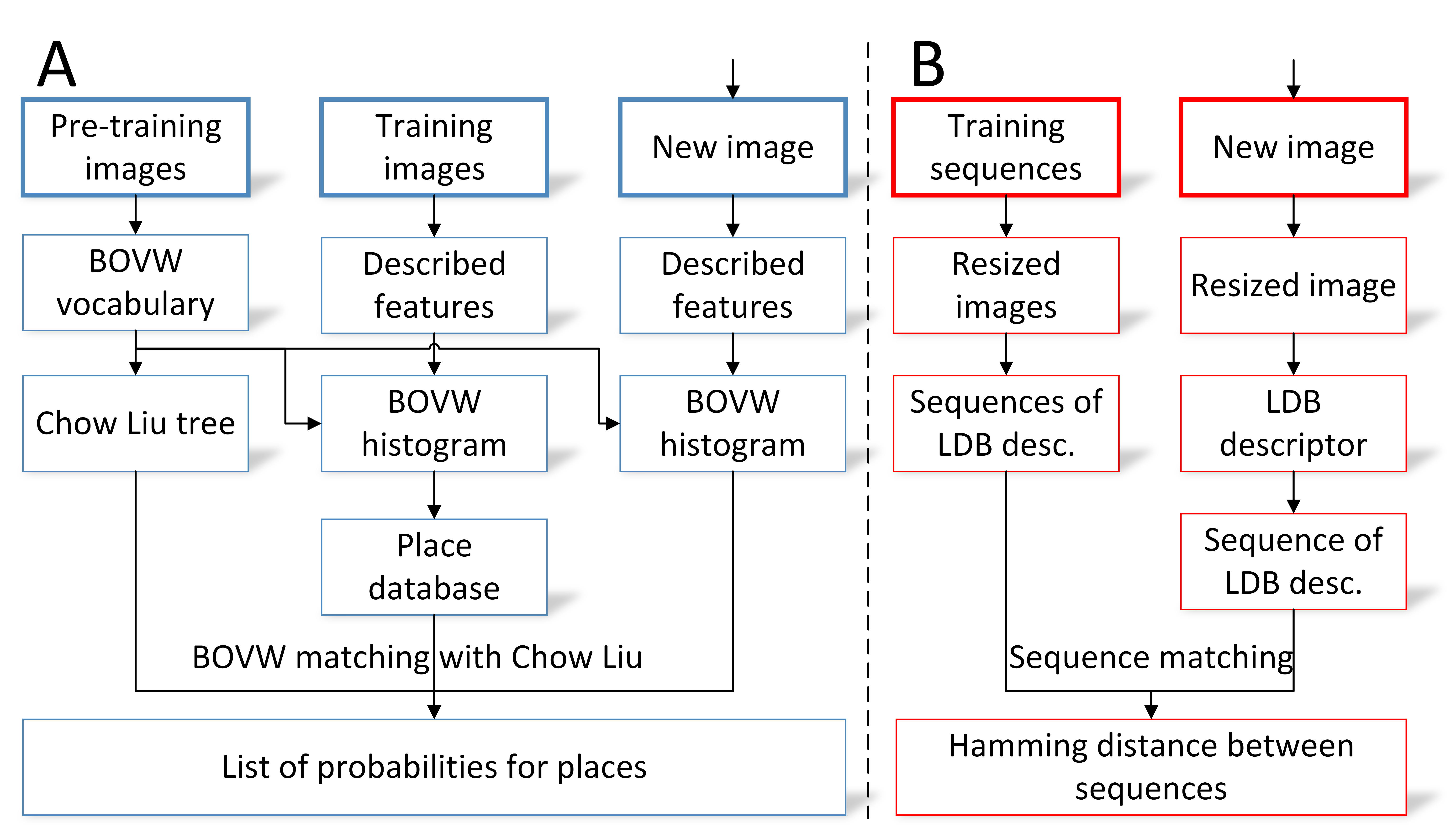}
 \caption{Block schemes of the FAB-MAP (A) and ABLE-M (B) algorithms}
\label{fig:scheme}
\end{figure}

During recognition, the FAB-MAP describes features in the image and computes the BOVW based on the occurrences of those descriptors.
The final step involves computing the set of probabilities that the observed image was taken in the place recorded in the database.
The $i$-th place is usually considered as recognized when the probability $P(L_i | \mathcal{Z}^k)$ exceeds the chosen threshold $t_p$.
The overview of the FAB-MAP processing is presented in Fig.~\ref{fig:scheme}a.

\subsection{Modifications to OpenFABMAP}
The OpenFABMAP system has one major parameter to tune, which is the probability threshold $t_p$.
In the preliminary tests \cite{IPIN}, we couldn't find a single setting of $t_p$ that would allow achieving no false positives
with the satisfactory number of correct recognitions. Therefore, we decided to modify the OpenFABMAP operation to include
information about a sequence of images. Whereas in the original OpenFABMAP the $i$-th place is
recognized in the $k$-th frame if the probability $P(L_i | \mathcal{Z}^k)$ exceeds $t_p$, in our improved version
the $i$-th place is considered to be recognized in the $k$-th frame only if the probability exceeds $t_p$ for all
frames in the window spanning from $(k-c_w+1)$-th to $k$-th frame:
\begin{equation}
    \forall_{j \in \{1,2,\ldots,c_w\}} P(L_i | \mathcal{Z}^{k - j +1}) > t_p,
\end{equation}
where $c_w$ is the size of the comparison window.
The introduced modification added the second parameter ($c_w$) to set up,
but finally allowed us to find a parametrization of OpenFABMAP that suits the indoor environment.
The value of $c_w$ influences a ratio between false positive and correct recognitions. The higher the value the longer
the comparison window and therefore the sequence of frames matched to the same image from the database has to be longer.
It implicates that higher values prevent the algorithm from making mistakes, but also lowers the number of places
that are correctly recognized. Analogously, a shorter $c_w$ window results in more false positives,
but also more places are recognized. Further discussion is provided in section \ref{sec:evalFabmap}.

\section{ABLE-M}
ABLE-M is the algorithm proposed by Arroyo et. al~\cite{able} that detects if the
camera observes a place that was already visited or determines if the currently
observed scene is similar to a previously recorded image. The ABLE-M algorithm is
available in the open-source implementation OpenABLE\footnote{\url{https://github.com/roberto-arroyo/OpenABLE}} \cite{openable}.
This implementation allows to process a single video file and determines if the given frame
(image) depicts a place that was already present in this sequence. This is typical loop closure
mode of operation, which is, however, impractical for personal indoor localization.
We want to determine the user location in real-time by comparing a short video sequence acquired on the go,
to a pre-recorded video of the environment. Therefore, our implementation of the system
operates on two sequences: the first one ($train$) is recorded prior to the localization
experiment and can be annotated with respect to the known floor plan of the building,
while the second one ($test$) is recorded on-line by the user and used as the query.

\subsection{Training and recognition}
The ABLE-M algorithm processes sequences of images, thus the processing of individual
frames has to be very fast to allow real-time operation. To this end, ABLE-M resizes
the processed images to the size of $64\times64$ pixels. Next, a global variant of
the LDB feature descriptor is computed with respect to the whole resized image.
The binary LDB descriptor~\cite{LDB} is computed by comparing individual intensities of
pairs of points along the predefined lines. LDB was designed to suit the needs of mobile devices,
thus it can be computed with minimal effort. It should be noted that prior to image description
ABLE-M can perform the reduction of illumination changes in the sequence using the method of
McManus {\em et al.} \cite{illu}. However, the illumination invariance is turned off by default
in the OpenABLE implementation. As illumination changes in building interiors are limited,
comparing it to outdoor scenes, we have decided to let this option be turned off, to make the processing faster.

The next step in the algorithm is the matching phase, in which the distance matrix
$\mathbf{D} = \{d_{i,j}\}$ between positions registered in the sequences $train$ and $test$ is computed.
As the $train$ sequence contains $n$ global image descriptors, and the $test$
sequence contains $m$ descriptors, the matrix size is $(n-c_l + 1)\times (m-c_l + 1)$:
\begin{equation}
    d_{i,j} = {\rm hamming}(train[i, i+c_l-1],test[j,j+c_l-1]),
\end{equation}
where $c_l$ is the length of the sub-sequence used for matching
(in~\cite{IPIN} it was denoted $compareLength$ and in~\cite{able} $d_{length}$).
The function ${\rm hamming}(x,y)$ computes the Hamming distance between the
descriptors of the sequences denoted as $x$ and $y$. Finally, the resulting
distance matrix $\mathbf{D}$ is normalized. The notation $train[i,i+c_l-1]$
represents the concatenation of descriptors from $i$-th to $(i+c_l-1)$-th frame of the $train$ sequence,

The computational complexity of the baseline algorithm (ABLE-M) is equal to $\mathcal{O}(n m c_l)$,
as for each image in the recognition sequence ($m$) we slide the window $n$
times and compare the sub-sequences in $c_l$ operations.

\subsection{Modifications made to OpenABLE}
The original ABLE-M algorithm compares two sequences of images to determine if loop closure has occurred.
The value in the distance matrix in the $i$-th row and $j$-th column corresponds
to the distance computed between the sub-sequence from $i$-th to $(i+c_l-1)$-th frame in the training sequence,
and the sub-sequence from $j$-th to $(j+c_l-1)$-th frame in the testing sequence. The $c_l$ parameter has to be chosen depending on the
environment characteristics. Originally, the distance matrix is normalized after the whole testing sequence
was processed and the places are marked as recognized if the value at the $(i,j)$-th position in the matrix
is lower than the recognition threshold $t_r$, set prior to operation. The $t_r$ value is the second parameter
that requires tuning depending on the environment. Both of parameters, $c_l$ and $t_r$, are set experimentally
and no formal learning phase is performed.

The choice of $c_l$ greatly impacts the performance of the algorithm.
Longer comparison windows make the system more robust, as larger parts of the training and testing trajectories
have to match to achieve a recognition from the system.
Unfortunately, longer windows also require the user to walk longer along the same trajectory as the one recorded
in training sequences, making the localization system impractical, whenever multiple crossroads are located
next to each other. Also, the OpenABLE system becomes more vulnerable to changes in the walking speed of the user,
as longer trajectories have to match each other. Shorter comparison windows, on the other hand, reduce the
recognition capabilities of the OpenABLE system, as less information is used for comparison,
which makes the system susceptible to local self-similarities.

Another problem with the practical application of OpenABLE for personal localization is the large number of
recognitions we usually obtain in our target environments. If the testing sequence contains a long trajectory
recorded along the same path as the training sequence, the system yields a large number of positive recognitions,
as the query sub-sequence of images can be successfully matched against multiple overlapping sub-sequences
in the training sequence. The value at $(i,j)$ position in the distance matrix is usually similar to
the value in $(i+1,j+1)$ position in the same matrix. This can be observed especially with larger values
of the window size $c_l$, as a large portion of the computed distance is the same in both of those cases.
The large number of recognitions in the same place inflates the statistics of positive matches,
but yet does not introduce a novel information about user location.

Therefore, we propose to cluster the positive recognitions obtained from OpenABLE to achieve a reasonable number
of distinct recognitions. In our implementation, we use DBScan~\cite{dbscan}, which is an
unsupervised clustering algorithm that does not require a priori information about the
expected number of clusters. In our modification of OpenABLE, the DBScan algorithm clusters
two-dimensional elements of the distance matrix ${\bf D}$. If both ${\bf D}(i,j)$ and ${\bf D}(k,l)$
exceeded the recognition threshold $t_r$, then those elements are clustered according to the Euclidean norm:
\begin{equation}
    dist({\bf D}(i,j),{\bf D}(k,l) ) = \sqrt{(i-k)^2 + (j-l)^2},
\end{equation}
which directly considers the indices of the distance matrix elements.
The algorithm is tuned according to two parameters -- the maximum distance $db_{\rm eps}$ that
allows adding a new point to the cluster, and the minimal number of points to form a cluster ($db_{\rm min\_pts}$).
The $db_{\rm eps}$ was set to 2, and $db_{\rm min\_pts}$ was equal to 1. To ease the comparison with FAB-MAP,
each cluster is represented by the average indices of the cluster members to verify the
correctness of the recognitions belonging to the cluster.

\section{Experimental evaluation}
The experimental evaluation of the OpenFABMAP and OpenABLE systems adopted to the
requirements of the personal localization task was conducted using images taken
with the Samsung Galaxy Note 3 smartphone with the Android operating system.
To increase the informational content of the images, they were taken while the
smartphone was held by the user in the horizontal position.
The resolution of images was set to $640 \times 480$. This resolution was chosen
as a compromise between the need to capture as much of the details in the environment as possible,
and the real-time processing and storage space constraints of the mobile device.
The choice of the resolution was also motivated by our previous experience with
the feature detectors and descriptors~\cite{iwcmc}, which are used in OpenFABMAP.
The ABLE-M algorithm processes images resized to $64 \times 64$ pixels,
thus the source image resolution influences it to a smaller extent. During the experiments,
the speed of capturing the images was limited to 8 frames per second due to the
limitations of the SD card in the smartphone.

\subsection{Experimental sites}
The experiments were conducted at Pozna\'n University of Technology (PUT) in two buildings
of quite a different topology. The PUT Lecturing Centre (PUT LC) is a modern style building with many
open spaces that are illuminated by sunlight coming trough multiple glass panels in the walls.
The PUT Mechatronic Centre (PUT MC) is a building of the more classic structure, with many
similar corridors. As the existing windows do not provide enough sunlight, artificial light is commonly used.

\begin{figure}[thpb!]
\centering
 \includegraphics[width=0.95\columnwidth]{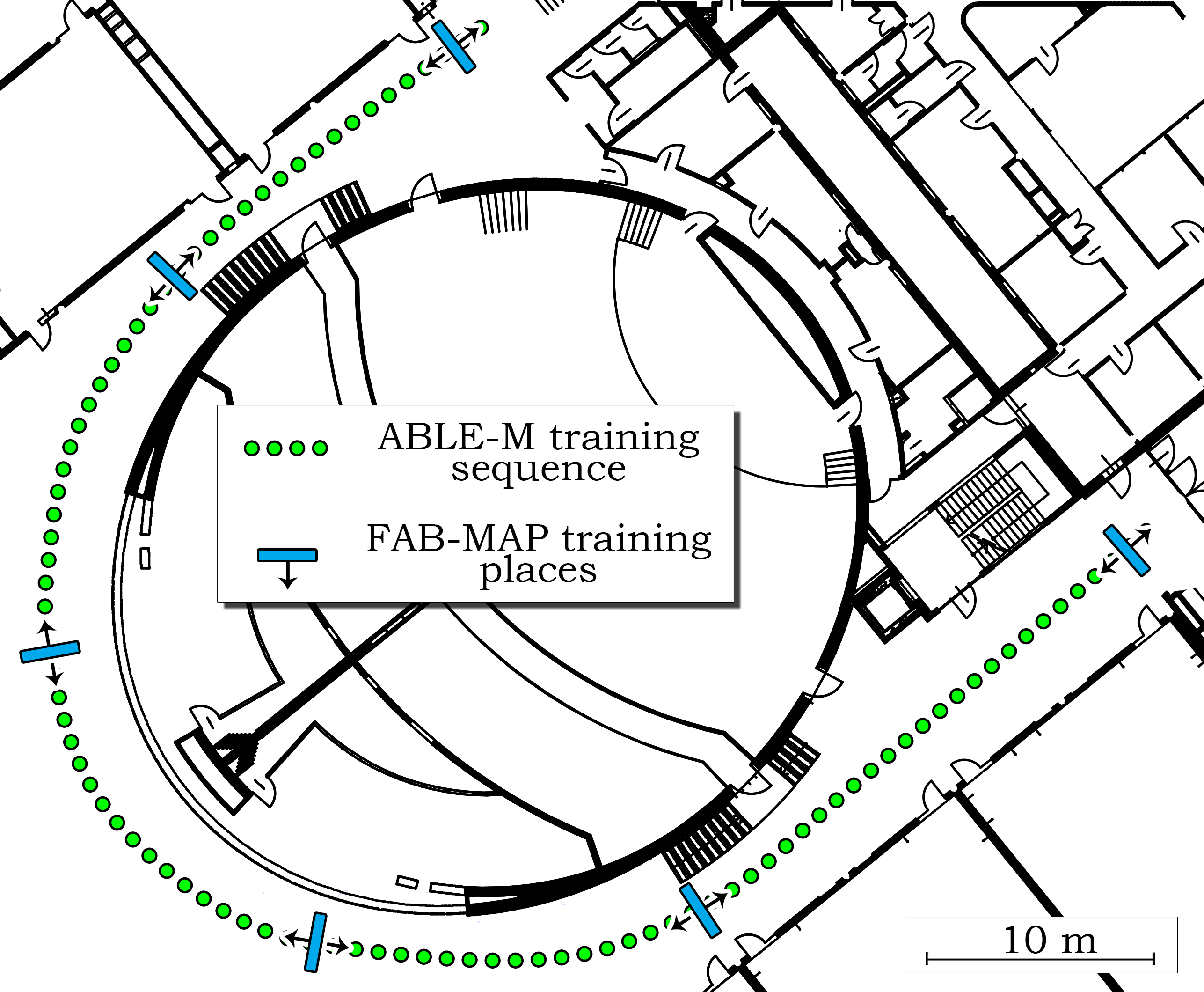}
 \caption{The PUT LC training sequence contains 6 recognition places with views
          in both directions for OpenFABMAP (blue rectangles with arrows) and 2
          sequences recorded in both directions for OpenABLE (green dots)}
\label{fig:CW}
\end{figure}

The first test inside the PUT LC building consisted of a single trajectory around the main lecturing hall.
During the training phase, 6 positions with images taken in both directions were taken for FAB-MAP,
and two sequences in both directions containing 458 and 438 images respectively were registered for ABLE-M.
In Fig.~\ref{fig:CW} the FAB-MAP training places are marked as blue rectangles,
whereas the ABLE-M training sequences are marked by a line of green dots. Additionally,
the training data were augmented with 5 images for FAB-MAP and a sequence of 410 images for ABLE-M
taken in the new PUT Library, which is a different building of a very similar structure to the PUT LC building.
The purpose of adding images from an outside of experimental sites was to make tests more challenging
and check if systems are capable of discriminating different sites.

After the training phase, the performance evaluation of the OpenFABMAP and OpenABLE systems was
accomplished on two test sequences acquired in the same environment: {\tt\scriptsize PUT LC test}
and {\tt\scriptsize PUT LC test 2}. The user was asked to walk roughly along the trajectory marked
by the green dots in Fig.~\ref{fig:CW}. The {\tt\scriptsize PUT LC test} sequence consists of 466 images,
and the {\tt\scriptsize PUT LC test 2} sequence contains 448 images.

\begin{figure}[thpb!]
\centering
 \includegraphics[width=\columnwidth]{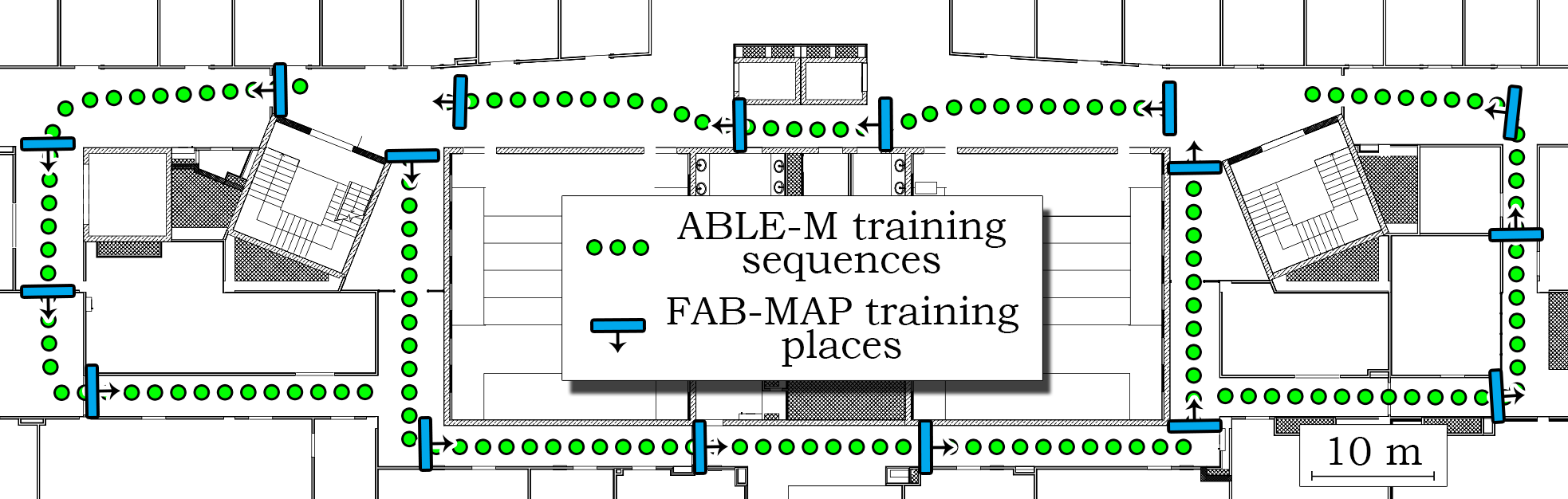}
 \caption{The PUT MC training sequence contains 17 recognition places for OpenFABMAP
 (blue rectangles with arrows) and 4 sequences for OpenABLE (green dots)}
\label{fig:CMtrain}
\end{figure}

The structure of PUT MC results in multiple junctions, and therefore the ABLE-M
learning trajectories can either map all available paths, or they can consist of
disjointed trajectories not covering the junctions. We have chosen the latter option
and divided PUT MC into 4 trajectories containing 1256 images altogether, as
presented by the green dots in Fig.~\ref{fig:CMtrain}. For the training purposes
of FAB-MAP 17 places were chosen, that are marked by blue rectangles in Fig.~\ref{fig:CMtrain}.
The final training data set for FAB-MAP was augmented by images of 5 places
from another floor of the same building, whereas the ABLE-M training data
additionally included one sequence of 274 images from that floor.

\begin{figure}[thpb!]
\centering
 \includegraphics[width=\columnwidth]{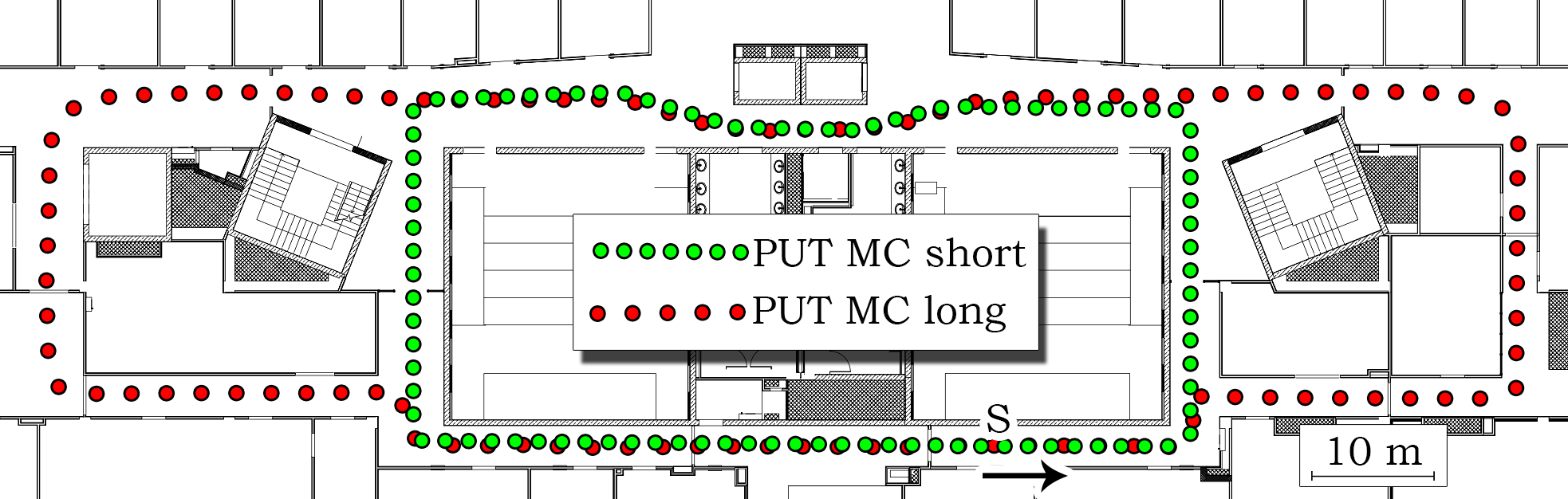}
 \caption{The trajectories of two PUT MC test sequences: {\tt\scriptsize PUT MC short},
          marked by orange dots, and {\tt\scriptsize PUT MC long}, marked by blue dashes}
\label{fig:CMtraj}
\end{figure}

Because PUT MC contains multiple corridors, we decided to have two different
test trajectories, {\tt\scriptsize PUT MC short} and {\tt\scriptsize PUT MC long},
recorded on the third floor. The trajectories are presented in Fig.~\ref{fig:CMtraj}
by green dots ({\tt\scriptsize PUT MC short}) and by red dots ({\tt\scriptsize PUT MC long}).
Both trajectories started and ended in a place marked by 'S' in Fig.~\ref{fig:CMtraj},
while the direction of motion is marked by an arrow. The shorter trajectory was equal
to approximately 60 metres and contains 620 images, while the longer trajectory is
approximately 120 metres long with 996 images. Taking into account the choice of
training places, 10 FAB-MAP test locations are put along the {\tt\scriptsize PUT MC short} trajectory,
and 15 test locations are put along the {\tt\scriptsize PUT MC long} trajectory.

To verify the robustness of the evaluated place recognition systems in populated environments,
we additionally recorded two sequences with people (individual or in groups) appearing at
random in the same corridors. These sequences were registered during a break between classes
when many students stay at the corridors or they wait at the doors of the classrooms.
The {\tt\scriptsize PUT MC people short} consists of 525 images, and in the {\tt\scriptsize PUT MC people long}
864 images were recorded.

\subsection{Evaluation procedure for OpenFABMAP}
\label{sec:evalFabmap}

The FAB-MAP algorithm needs a pre-training and training process.
For the pre-training we used a separate set of images that were collected in the same environment,
was not overlapping with the database of known places used for training. 
As it comes to the parameters of the modified OpenFABMAP, they need to be adjusted before the
system can be applied in real-life operation. The {\tt\scriptsize PUT LC test} and {\tt\scriptsize PUT MC short}
were chosen for the initial validation and the results of system testing are presented
in Tab.~\ref{tab:fabmapResVal}. As already mentioned, setting the sequence size $n$ to 1
results in incorrect recognitions even if the recognition threshold $t_p$ is set to 0.99.
As any incorrect recognition may be critical for indoor localization~\cite{mobicase},
these results strongly suggest that OpenFABMAP in the original version was incapable of
operation in the target environment. Increasing $c_w$ allows to suppress the number of incorrect
recognitions to 0, but might also lower the number of correct recognitions as the system
needs to properly match several images in a sequence. Therefore, the window size of 3 or 5
is the best choice in our case. As performing recognition on several images is more descriptive,
it is possible to lower $t_p$ increasing the number of correct recognitions without any additional incorrect matches.

\begin{table}[htpb!]
    \caption{Testing various OpenFABMAP parameters on validation sequences {\tt\scriptsize PUT MC short} and {\tt\scriptsize PUT LC test}}
    \centering
    \begin{tabular}{lrlrrrr}
        & & & \multicolumn{4}{c}{comparison window $c_w$} \\ \cline{4-7}
        sequence & \multicolumn{1}{l}{$t_p$ }& recognitions & 1 & 3 & 5 & 8 \\ \hline
        \multirow{6}{*}{{\tt\scriptsize PUT MC short}} & \multirow{3}{*}{0.6} & correct & 241 & 136 & 92 & 67 \\
        & & incorrect & 36 & 0 & 0 & 0 \\
        & & distinct & 9 & 9 & 6 & 5 \\ \cline{2-7}
        & \multirow{3}{*}{0.99} & correct & 174 & 85 & 59 & 38 \\
        & & incorrect & 1 & 0 & 0 & 0 \\
        & & distinct & 9 & 6 & 5 & 4 \\ \hline
        \multirow{6}{*}{{\tt\scriptsize PUT LC test}} & \multirow{3}{*}{0.6} & correct & 223 & 140 & 94 & 63 \\
        & & incorrect & 207 & 23 & 2 & 0 \\
        & & distinct & 6 & 6 & 5 & 4 \\ \cline{2-7}
        & \multirow{3}{*}{0.99} & correct & 202 & 126 & 94 & 65 \\
        & & incorrect & 122 & 11 & 1 & 0  \\
        & & distinct & 6 & 6 & 5 & 4 \\ \hline
    \end{tabular}
     \label{tab:fabmapResVal}
\end{table}

\begin{table}[htpb!]
    \caption{OpenFABMAP results in PUT MC and PUT LC with $t_p = 0.99$ for $c_w = 1$ and $t_p = 0.6$ for $c_w = 3, 5, 8$}
    \centering
    \begin{tabular}{llrrrr}
        & & \multicolumn{4}{c}{sequence length $n$} \\ \cline{3-6}
        sequence & recognitions & 1 & 3 & 5 & 8 \\ \hline
        \multirow{3}{*}{{\tt\scriptsize PUT LC test 2}} & correct & 224 & 199 & 168 & 135 \\
        & incorrect & 52 & 6 & 1 & 0 \\
        & distinct & 6 & 6 & 6 & 6 \\ \hline
        \multirow{3}{*}{{\tt\scriptsize PUT MC long}} & correct & 301 & 229 & 173 & 119 \\
        & incorrect & 22 & 2 & 0 & 0 \\
        & distinct & 13 & 12 & 10 & 7 \\ \hline
        \multirow{3}{*}{{\tt\scriptsize PUT MC people long}} & correct & 150 & 84 & 42 & 15 \\
        & incorrect & 24 & 3 & 0 & 0 \\
        & distinct & 10 & 7 & 7 & 5 \\ \hline
        \multirow{3}{*}{{\tt\scriptsize PUT MC people short}} & correct & 34 & 18 & 10 & 2 \\
        & incorrect & 4 & 0 & 0 & 0 \\
        & distinct & 9 & 4 & 3 & 2 \\ \hline
    \end{tabular}
    \label{tab:fabmapRes}
\end{table}

\begin{figure}[thpb!]
\centering
 \includegraphics[width=\columnwidth]{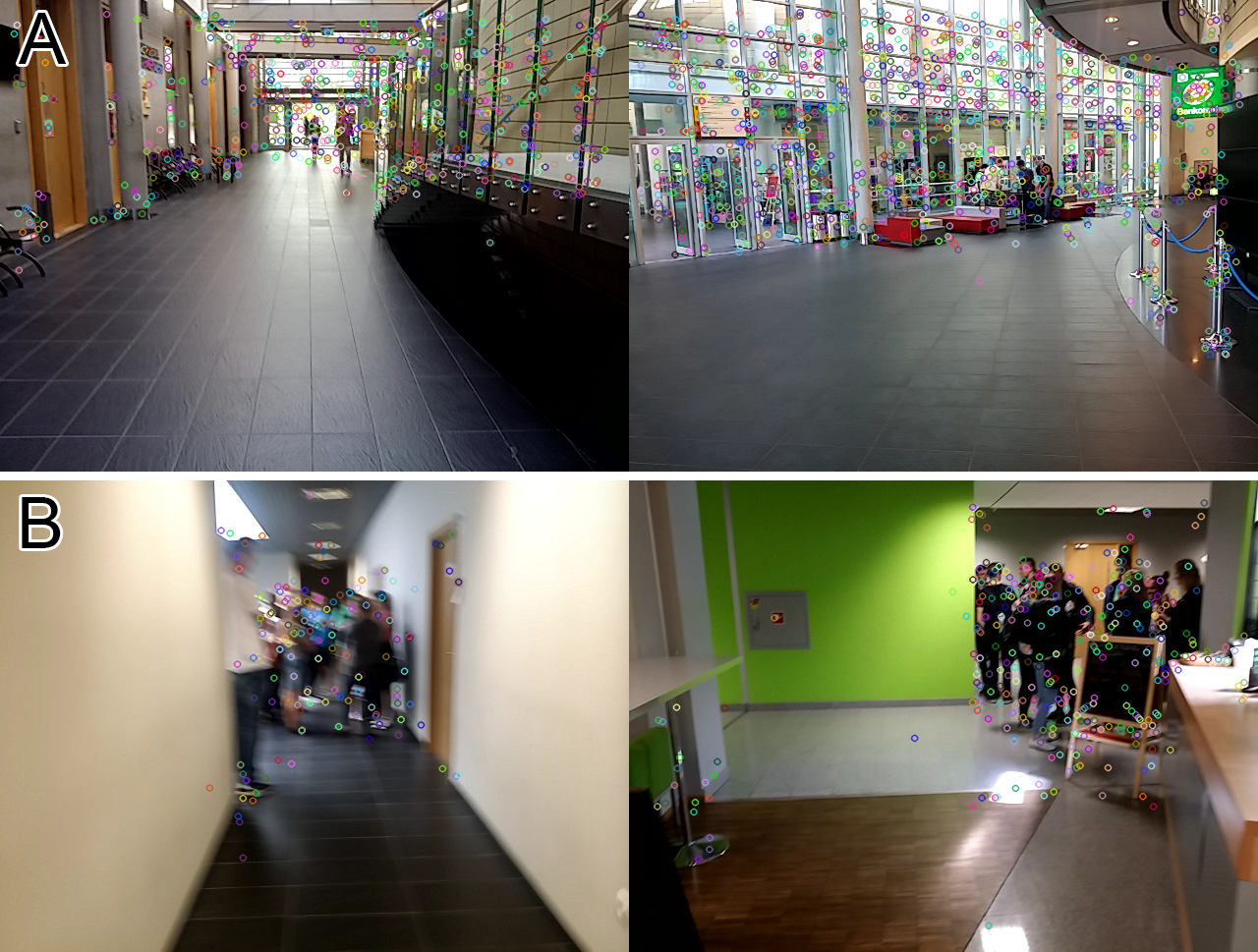}
 \caption{OpenFABMAP incorrectly recognized locations: A) due to many features placed on windows, B) due to people occluding the view}
\label{fig:fabmapFalsePositive}
\end{figure}

The parameters found in the previous analysis were used on datasets {\tt\scriptsize PUT LC test 2} and {\tt\scriptsize PUT MC long},
and the datasets with people moving inside corridors {\tt\scriptsize PUT MC people long} and {\tt\scriptsize PUT MC people short}.
The results confirm that the proper choice of windows size $c_w$ usually allows minimizing the number of incorrect matches.
In some cases, the characteristic of the environment contains multiple, misleading features making recognition a challenging task.
Such case is presented in Fig.~\ref{fig:fabmapFalsePositive}A as multiple features are visible on windows, misleading the FAB-MAP.
Similarly, the people present in the corridors make it difficult to correctly recognize the places. Such situation is presented in Fig.~\ref{fig:fabmapFalsePositive}B.


\subsection{Evaluation procedure for OpenABLE}
The OpenABLE results depend on the choice of two parameters. The first one, $c_l$ denotes the
number of frames used when comparing sequences whereas the second $t_r$ is the recognition threshold
in the normalized distance matrix. The authors of OpenABLE do not provide any procedure for setting
those parameters, and therefore we perform an experimental evaluation for different settings on {\tt\scriptsize PUT LC test} dataset.
The most significant results are presented in Tab.~\ref{tab:ableCWRes} containing tests for $c_l$
ranging from 2 to 40 images and two setting of $t_r$. For the convenience of the readers, we only
provide the clustered recognized places, which is a more intuitive performance measure than the
total number of positive results from OpenABLE.

\begin{table}[htpb!]
 \caption{Influence of parameters $c_l$ and $t_r$ in OpenABLE on {\tt\scriptsize PUT LC test}}
    \centering
    \begin{tabular}{llrrrrr}
        &  &  \multicolumn{5}{c}{$c_l$}  \\ \cline{3-7}
        \multirow{2}{*}{$t_r$} & recognized & \multirow{2}{*}{2} & \multirow{2}{*}{5} &\multirow{2}{*}{10} & \multirow{2}{*}{20} & \multirow{2}{*}{40}\\
         & places &  &  &  &  & \\ \hline
        \multirow{2}{*}{0.4} & correctly & 61 & 36 & 19 & 21 & 6\\
        & incorrectly & 387 & 105 & 42 & 4 & 0\\ 
        \hline
        \multirow{2}{*}{0.3} & correctly & 121 & 49 & 16 & 15 & 8\\
        & incorrectly & 57 & 4 & 0 & 0 & 0\\ 
        \hline
    \end{tabular}
     \label{tab:ableCWRes}
\end{table}

The OpenABLE algorithm was at first tested with $t_r$ set to 0.4 and different sizes of $c_l$.
For small window sizes, likes 2 or 10, the algorithm recognizes more incorrect than correct
places but even for $c_l$ equal to 2 the number of positive hits from OpenABLE (16936) exceeds
the number of incorrect recognitions (10962). The poor performance can be attributed to the fact
that OpenABLE extracts not enough information from individual images, as those images are
reduced in size and described by a global descriptor.
The strength of the algorithm lies in exploring the sequences of images. For greater $c_l$
the system recognizes places, but it also negatively affects the number of positive recognitions
as longer sequences have to match in order to obtain a correct place recognition.

Similar results can be also obtained for $t_r$ set to 0.3. In this case, OpenABLE achieves
fewer incorrect matches due to the fact that $t_r$ is a threshold that is more strict.
Regardless of the choice of $t_r$ to 0.3 or 0.4, the OpenABLE results in 0 incorrect recognitions
when comparison window $c_l$ is set to 40.
The sequence of 40 images in our case corresponds to the operation for approximately 5 seconds.
From our experience, the value of $t_r$ is influenced by the structure of images and usually is
hard to tune for sequences containing different parts of the building and the value of $c_l$
should be chosen based on the similarity of places in the dataset. For example, if the database
consists of several similar corridors, it is usually better to increase $c_l$.
From the first experiment, we assumed $c_l$ to be 40 and $t_r$ of 0.4 and wanted to verify
the performance inside the building with a different structure.
The choice of those parameters was also evaluated on {\tt\scriptsize PUT LC test 2}.
The system correctly recognized 6 places on the trajectory with 4529 positive recognitions from OpenABLE.
Moreover, the system presented no false positives.

\begin{table}[htpb!]
    \caption{The results obtained with OpenABLE for different $c_l$ on datasets from PUT MC}
    \centering
    \begin{tabular}{llrrr}
        & Recognized & \multicolumn{3}{c}{$c_l$} \\ \cline{3-5}
        Sequence & places & 40 & 60 & 80 \\ \hline
        {\tt\scriptsize PUT MC short} & correctly & 10 & 6 & 5\\
        ($t_r=0.4$) & incorrectly & 6 & 1 & 0 \\ 
        \hline
        {\tt\scriptsize PUT MC long} & correctly & 13 & 10 & 8 \\
        ($t_r=0.4$) & incorrectly & 8 & 0 & 0 \\ 
        \hline
        {\tt\scriptsize PUT MC people short}  & correctly & 5 & 3 & 1\\
        ($t_r=0.4$) & incorrectly & 1 & 0 & 0 \\ 
        \hline
        {\tt\scriptsize PUT MC people long} & correctly & 8 & 1 & 0\\
         ($t_r=0.4$) & incorrectly & 0 & 0 & 0 \\ 
        \hline
        {\tt\scriptsize PUT MC people short}  & correctly & 9 & 8 & 2\\
        ($t_r=0.5$)  & incorrectly & 33 & 5 & 1 \\ 
        \hline
        {\tt\scriptsize PUT MC people long} & correctly & 12 & 11 & 2\\
        ($t_r=0.5$) & incorrectly & 24 & 0 & 0 \\ 
        \hline
    \end{tabular}
    \label{tab:ableCMRes}
\end{table}

The PUT MC has a different structure than PUT LC and therefore the parameters of
OpenABLE have to take into account those differences. The results obtained on
{\tt\scriptsize PUT MC short}, {\tt\scriptsize PUT MC long} and the sequences
with people being present in the field of view are shown in Tab.~\ref{tab:ableCMRes}.

\begin{figure}[thpb!]
\centering
 \includegraphics[width=0.9\columnwidth]{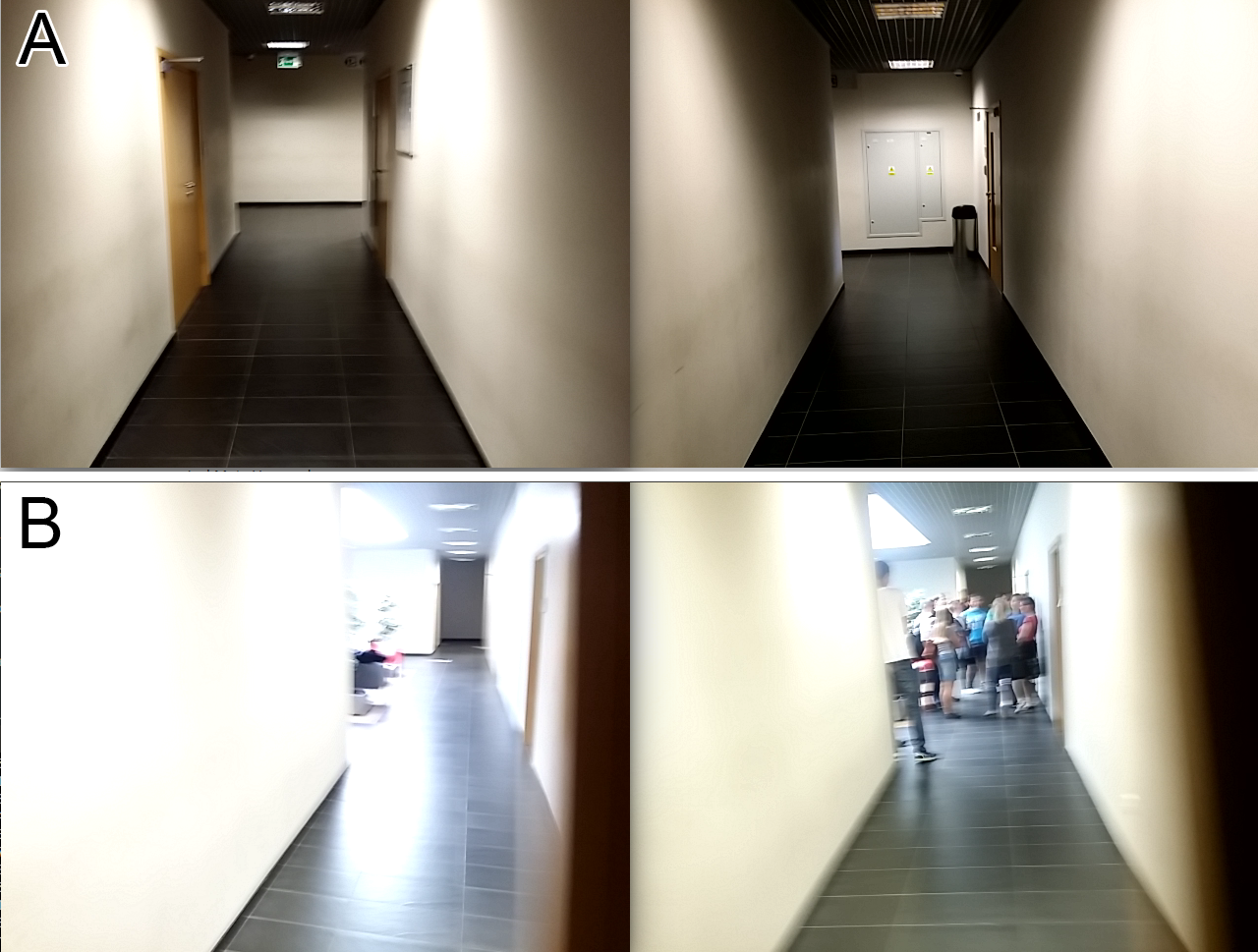}
 \caption{{\tt\scriptsize PUT MC} is a challenging environment due to
          similarities that can cause incorrect recognitions for
          ABLE-M (A). But in some real-life cases ABLE-M correctly recognizes
          locations in corridors even with multiple people visible in images (B)}
\label{fig:ablePeople}
\end{figure}

Setting the $c_l$ results in almost every case in some incorrectly recognized places.
Even if there are no incorrectly recognized places for $c_l$ equal to 40, small change
in $t_r$ from 0.4 to 0.5 results in significant increase of incorrect recognitions,
which can be observed for {\tt\scriptsize PUT MC short} and {\tt\scriptsize PUT MC long}.
As corridors are hard to distinguish even for humans, the window size $c_l$ was extended
as shorter sequences might not provide sufficient information for place recognitions.
This is necessary as building structure contains similar corridors, like those presented in Fig.~\ref{fig:ablePeople}A.
Increasing $c_l$ to 80 images corresponds to the 8 seconds of motion. 

Another important aspect of real-life operation is the influence of people on recognition. 
Therefore, we recorded and can compare the same trajectories with and without people
occluding the view from the camera. In both cases, {\tt\scriptsize PUT MC people short}
and {\tt\scriptsize PUT MC people long}, there were no incorrect recognitions when $c_l$ was set to 60.
Compared to sequences without people, the system recognizes fewer different places.
This is normal as people can occlude a significant part of the images changing the properties of
images for comparison. Therefore, large values of $c_l$ might provide more robust recognitions
in the case of small disturbance, like 1 person, but not be efficient in case of a crowd as the
system compared longer trajectories. Nevertheless, OpenABLE exhibits impressive performance
correctly recognizing places even in such situations as the one presented in Fig.~\ref{fig:ablePeople}B.

\subsection{Performance comparison between OpenFABMAP and OpenABLE}
The initial experiments proved that OpenFABMAP and OpenABLE can be successfully used in indoor environments.
Nevertheless, the obtained results cannot be simply compared with respect to the number of correct
and incorrect recognitions, due to the conceptual differences between the FAB-MAP and ABLE-M algorithms.
Therefore, we present the results on shared figures that allow to visually compare the achieved recognitions
and to draw conclusions about the algorithms. In each case, we present the best results for parameter
settings that yielded no incorrect place recognitions.

\begin{figure}[thpb!]
\centering
 \includegraphics[width=0.85\columnwidth]{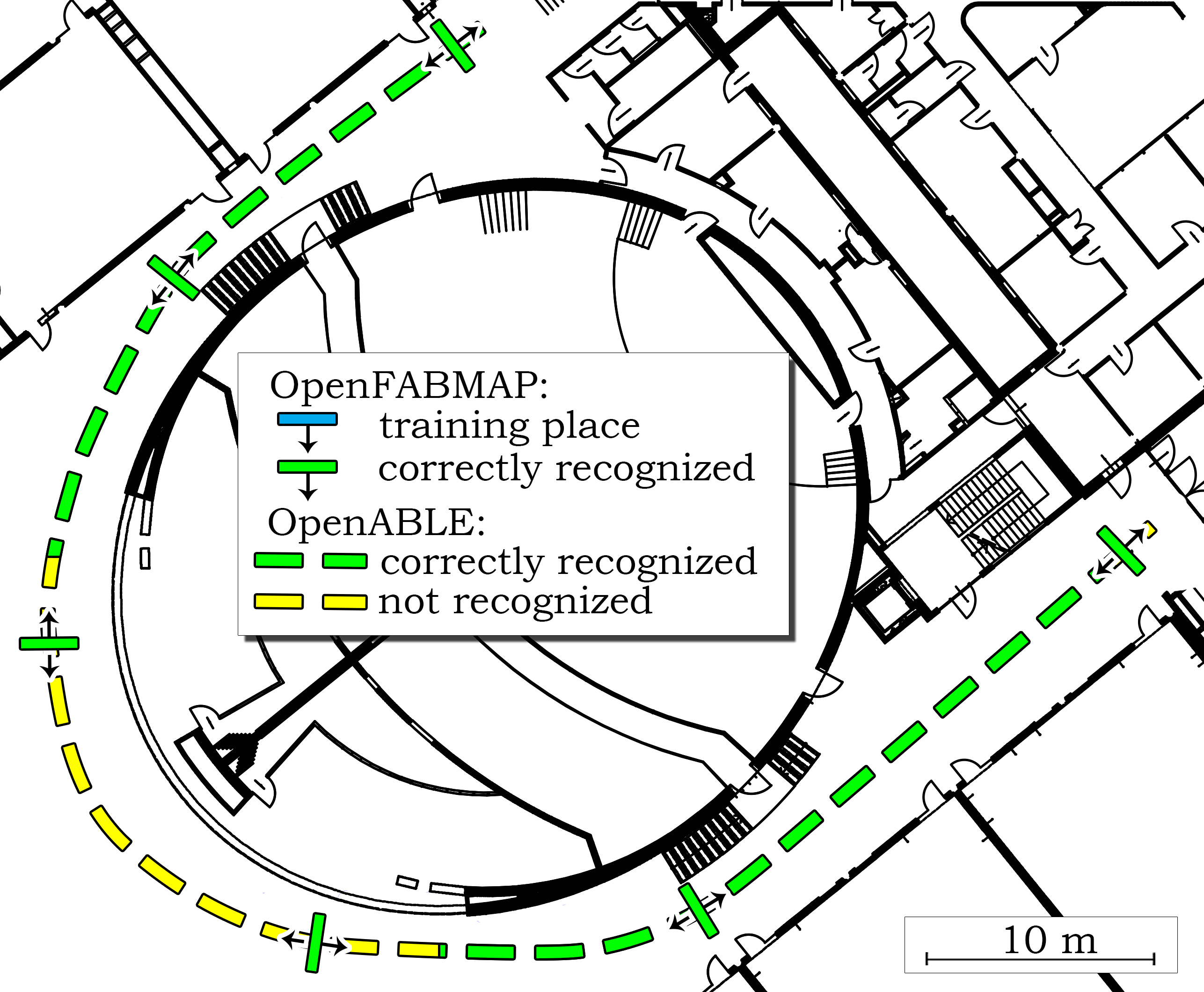}
 \caption{The results obtained with the OpenFABMAP and OpenABLE on {\tt\scriptsize PUT LC test 2}}
\label{fig:compareCW2}
\end{figure}

In Fig.~\ref{fig:compareCW2} we present the results obtained on {\tt\scriptsize PUT LC test 2} sequence.
It can be observed that the OpenFABMAP correctly recognized all of the places along the test trajectory.
The OpenABLE failed to achieve recognitions in the middle part of the trajectory but provided continuous
localization in the beginning and the end of the sequence. Depending on the application, both systems
provided valuable information for the localization purposes.

\begin{figure}[thpb!]
\centering
 \includegraphics[width=\columnwidth]{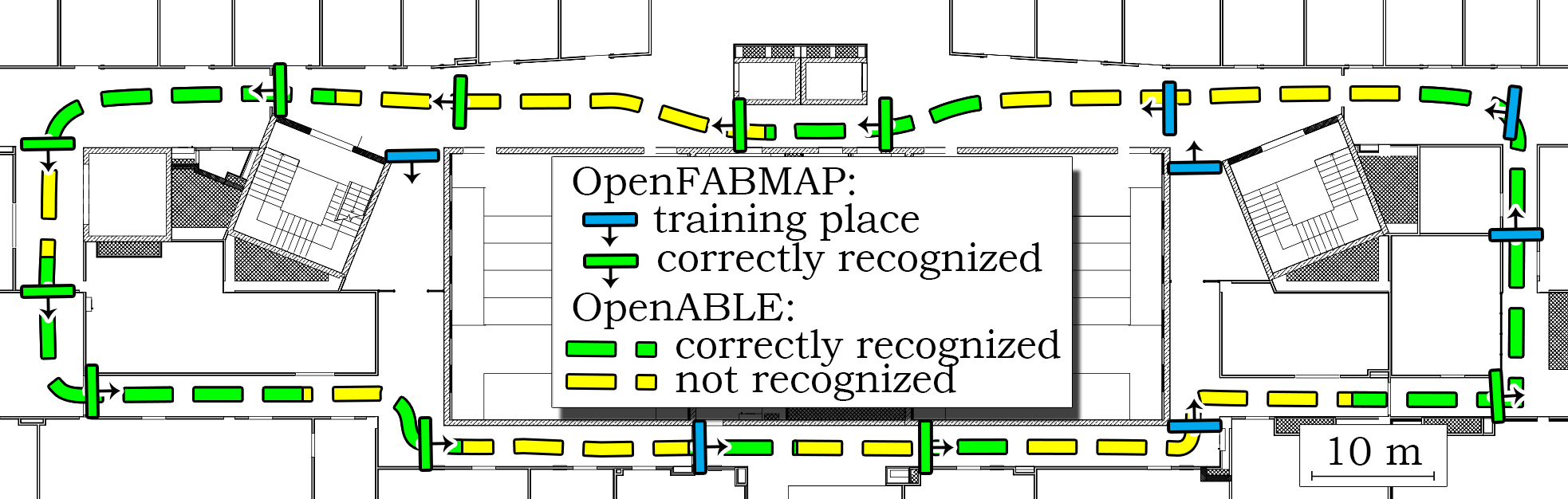}
 \caption{The results obtained with the OpenFABMAP and OpenABLE on {\tt\scriptsize PUT MC long}}
\label{fig:compareFABMAPABLE}
\end{figure}

In Fig.~\ref{fig:compareFABMAPABLE} we present the results obtained for the {\tt\scriptsize PUT MC long} sequence.
The OpenFABMAP managed to correctly recognize user location on 10 out of 15 possible places along the trajectory.
The OpenABLE recognized the user in 7 parts of the trajectory, but some of those parts are equal to several
seconds providing the system with continuous user localization. The main difference between algorithms is visible close to junctions --
OpenABLE cannot recognize user on those locations, while OpenFABMAP works correctly. This suggests that these
solutions are suitable for buildings of different structure or might be combined in a robust visual place recognition system.

\begin{figure}[thpb!]
\centering
 \includegraphics[width=\columnwidth]{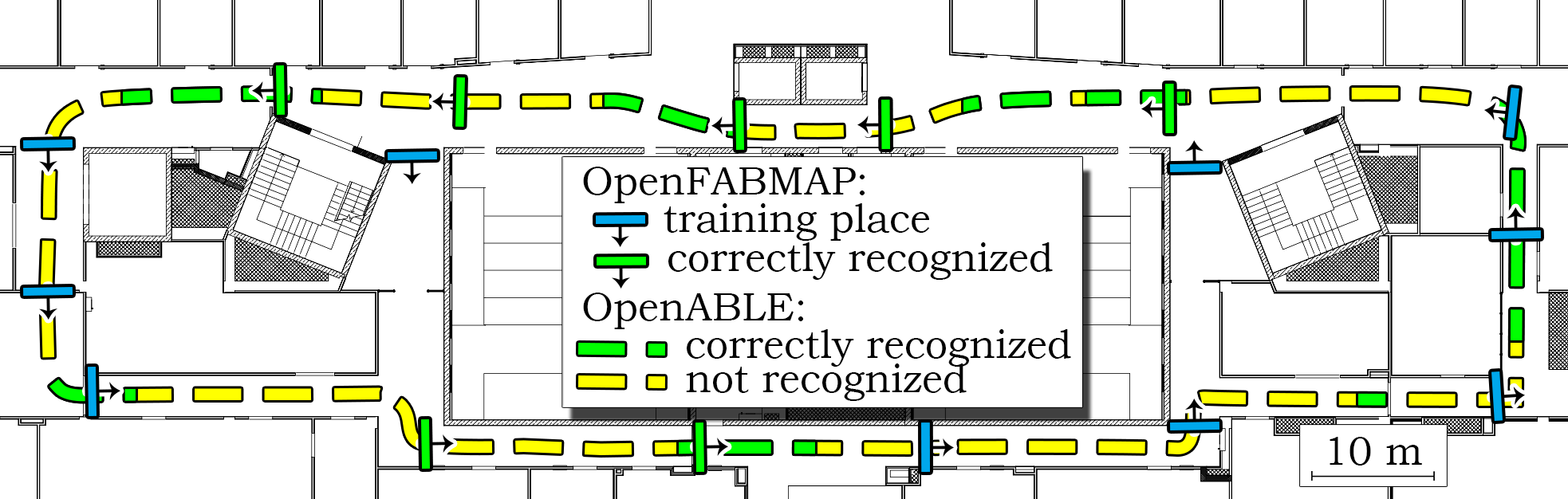}
 \caption{The results obtained with the OpenFABMAP and OpenABLE on {\tt\scriptsize PUT MC people long} containing additional people compared to {\tt\scriptsize PUT MC long}}
\label{fig:comparepeople}
\end{figure}

One of the main challenges of visual place recognition is the presence of people in front of the camera.
Therefore, in Fig.~\ref{fig:comparepeople} we present the results obtained along the same trajectory as
{\tt\scriptsize PUT MC long}, but during the student break with many people walking along the corridors.
In such conditions, the OpenFABMAP recognized the user in 7 out of 15 possible places along the trajectory.
The OpenABLE recognized the user in 7 parts of the trajectory, but the recognized parts were shorter when compared
to the recognition results obtained when the people were not present.
Both systems correctly recognized the images of re-visited places if the presence od people only slightly disturbed
the original view of the scene. Therefore, in many cases, the OpenFABMAP was able to correctly recognize a place
based on images taken when people just passed by, which was not possible to OpenABLE.
On the other hand, OpenABLE required a longer sequence of similar images for successful recognition
but was more robust to small, but continuous disturbance introduced by the presence of people,
which led to some surprisingly correct recognitions, as presented in Fig.~\ref{fig:ablePeople}B.
Neither of the algorithms provided incorrect recognitions due to the people obscuring the field of view.


\section{Processing time analysis}
The processing time is an important factor often deciding if the algorithm can be
applied on a mobile device in real-life scenarios. Moreover, it is related to the energy consumption,
as we are not only interested in localization systems that run in real-time,
but also in solutions that do not require the whole processing time of the mobile device's CPU.

To verify the processing requirements, we implemented OpenFABMAP and OpenABLE systems on
Samsung Galaxy Note 3 and Lenovo S8-50, hereinafter abbreviated to SGN3 and LS8-50.
Both of those devices are equipped with Android 4.4. Systems are implemented in C++ and
therefore we used C++ NDK to create shared libraries called from the Android application written in Java.
For image processing, we used OpenCV library in version 3.2.0 with a contrib module for SURF and SIFT.
Compared to our preliminary results~\cite{IPIN}, the applications were moved from the Eclipse IDE
to the Android Studio due to the depreciation of Eclipse IDE and the lack of support from Google
in new versions of Android. Therefore, the run times measured in the new versions of our
applications are different than in \cite{IPIN}, due to differences in the compiler optimization.

\begin{figure}[thpb!]
\centering
 \includegraphics[width=\columnwidth]{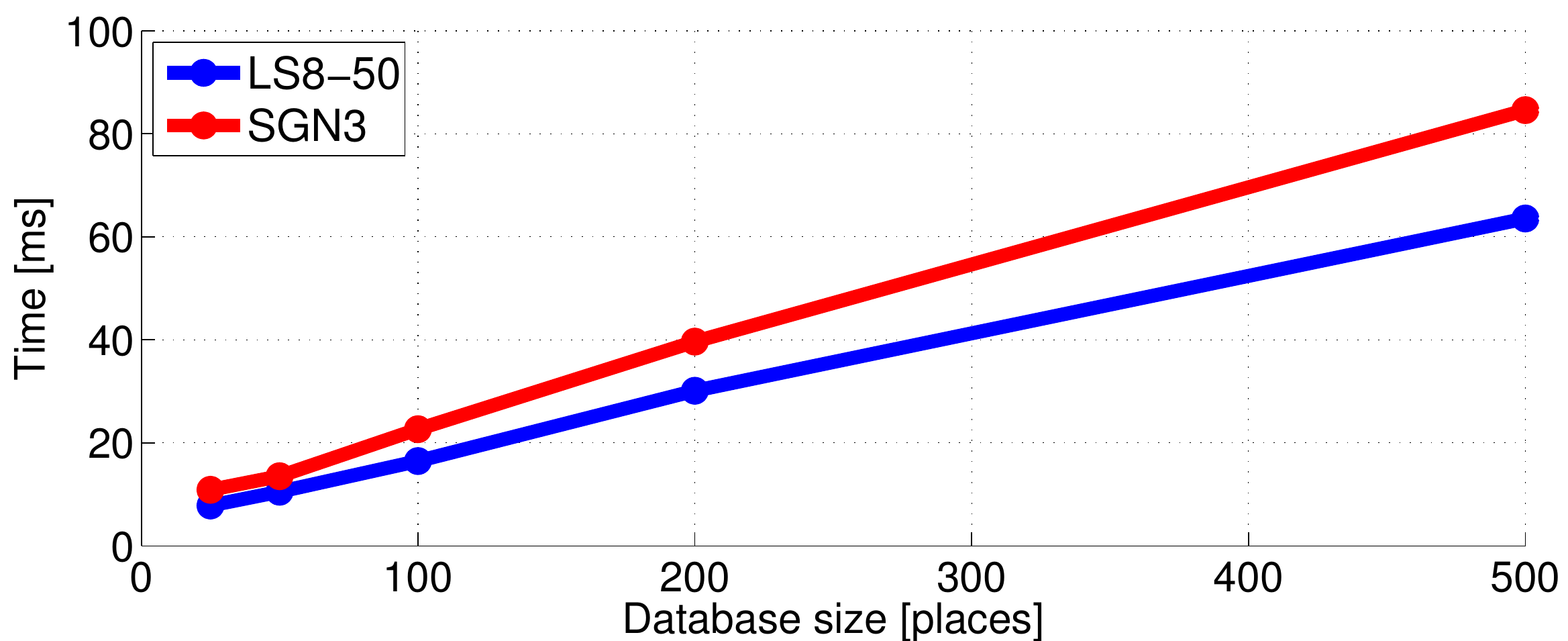}
 \caption{The image to database matching time for OpenFABMAP algorithm depending on the number of places in the database for Lenovo S8-50 (LS8-50) and Samsung Galaxy Note 3 (SGN3)}
\label{fig:timefabmap}
\end{figure}

The OpenFABMAP processing time can be divided into a single image processing time and a place recognition time.
The single image processing time is the most time-consuming part as extraction of SURF features takes 156 ms
per frame on SGN3 and 92 ms on LS8-50. The description of features is also time demanding as it takes exactly 89 ms per frame on both devices.
Therefore, single frame processing without matching to the database takes around 245 ms on SGN3 and 181 ms on LS8-50.
The place recognition time depends on the size of the database and the obtained results are presented in Fig.~\ref{fig:timefabmap}.
In those experiments, the time needed to match the processed image to the database was equal to several milliseconds,
growing linearly with the number of places stored in the database.

\begin{figure}[thpb!]
\centering
 \includegraphics[width=\columnwidth]{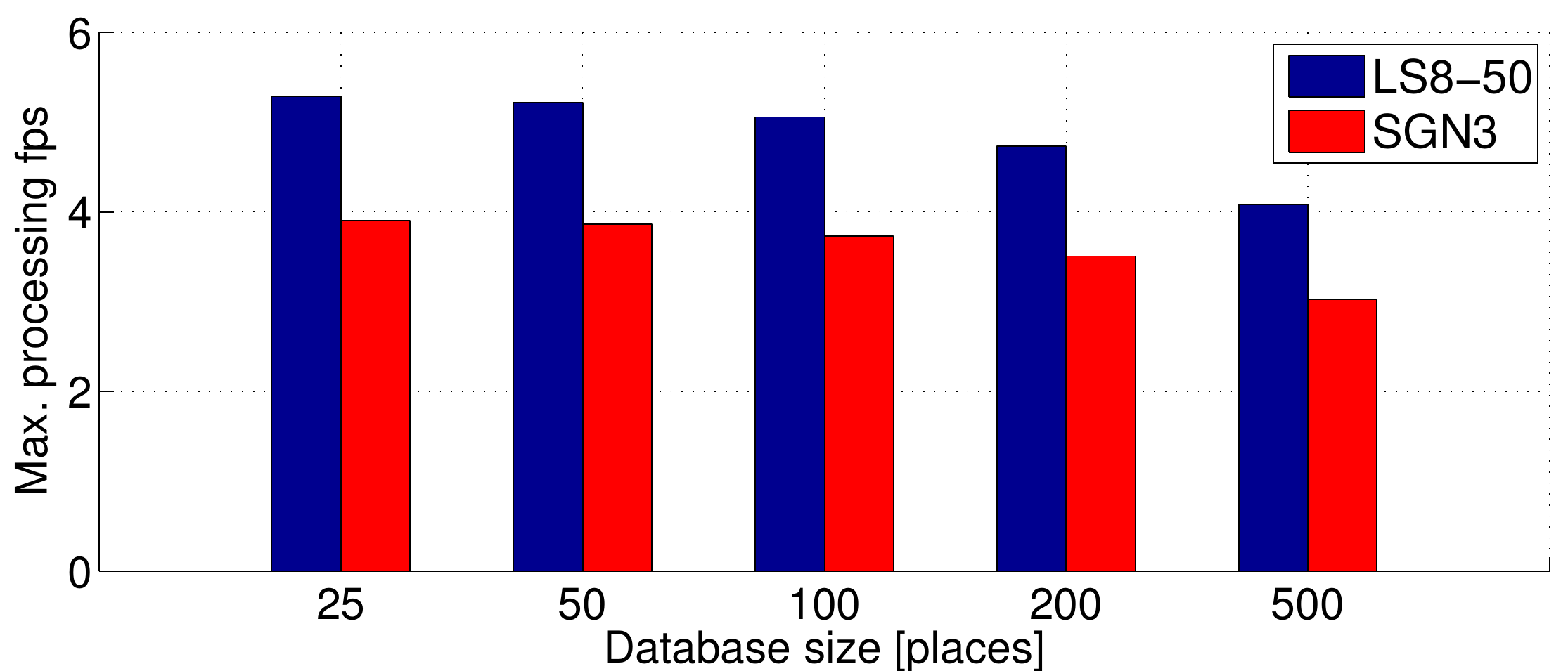}
 \caption{Maximum number of processed frames per second for OpenFABMAP depending on the database size for Samsung Galaxy Note 3 (SGN3) and Lenovo S8-50 (LS8-50)}
\label{fig:timefabmapfps}
\end{figure}

The maximum numbers of processed frames per second for different databases are presented in Fig.~\ref{fig:timefabmapfps}.
In our experiments, the databases contained no more than 50 frames but even for larger databases the maximal processing
frames per second are almost the same. Unfortunately, the OpenFABMAP system can only process around 5 frames per second
for SGN3 and LS8-50, which is insufficient in the case of experiments in PUT buildings. Even if the phone would record
images with the desired framerate, the processing would take $100\%$ CPU time, which will have a negative
impact on the battery life of a mobile device.

\begin{figure}[htbp!]
 \includegraphics[width=\columnwidth]{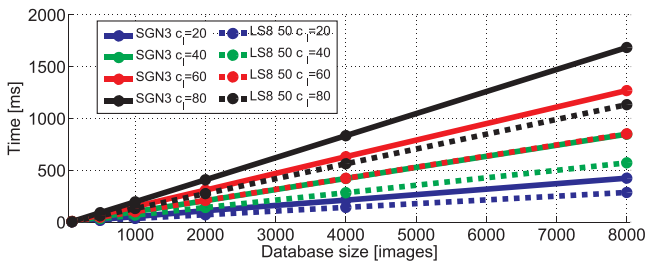}
 \caption{Time taken to match a new sequence of images to the pre-recorded database
          depending on the database size and the length of comparison window for OpenABLE on Samsung Galaxy Note 3 (SGN3) and Lenovo S8-50 (LS8-50)}
 \label{fig:ableOldTime}
\end{figure}

Similarly to the OpenFABMAP, OpenABLE processing time can be divided into single image processing and sequence matching times.
Single image processing time for OpenABLE is almost negligible as image resizing takes 0.48 ms per frame and computing LDB descriptor takes 0.44 ms per frame.
Thus the total time needed to process single image is approximately equal to 1 ms.
However, the sequence matching is the time-consuming part. 
The processing time of OpenABLE depends on the length of the sequence $c_l$ and the size of the database.
We assume that the video data should be processed at 10 frames per second to achieve real-time localization.
Therefore, the sizes of the database varying from 100 to 4000 images represent camera trajectories
recorded during motion lasting from 10 to 400 seconds. The size of the comparison window
was also chosen experimentally and was varying from 20 to 80 frames.
Results of experiments with several different settings for those parameters are
presented in~Fig.~\ref{fig:ableOldTime}. During those experiments, the testing sequence was equal to 1000 frames to average obtained times.

\begin{figure}[htbp!]
 \includegraphics[width=\columnwidth]{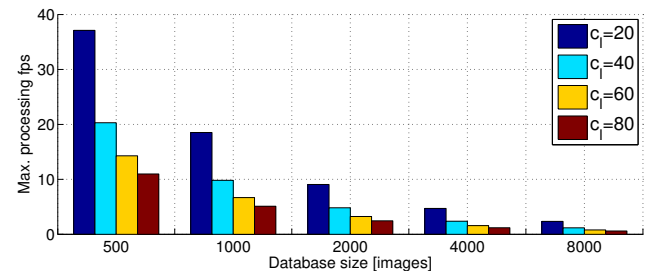}
 \caption{Maximum number of processed frames per second for OpenABLE
          depending on the database size and comparison window for Samsung Galaxy Note 3 (SGN3)}
 \label{fig:cpuTimeSGN3}
\end{figure}

\begin{figure}[htbp!]
 \includegraphics[width=\columnwidth]{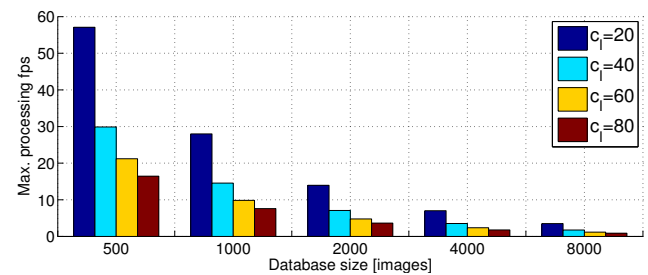}
 \caption{Maximum number of processed frames per second in OpenABLE depending
          on the database size and comparison window for Lenovo S8-50}
 \label{fig:cpuTimeLenovo}
\end{figure}

From the plots in Fig.~\ref{fig:ableOldTime} it is evident that OpenABLE can be only applied
in environments of limited size, as the matching times increase linearly not only with the
growing size of the database but also with the length of the comparison window.
The total processing time reaches 832 ms for SGN3 when $c_l$ is equal to 80
and the database size contains 4000 images. The time obtained for LS8-50 for
the same parameters is equal to 562 ms. Matching times over 125 ms prevent the
system from operating in real-time. The exact number of frames per second that
can be processed in real-time on Samsung Galaxy Note 3 is presented in
Fig.~\ref{fig:cpuTimeSGN3} whereas the results for Lenovo S8-50 are presented in Fig.~\ref{fig:cpuTimeLenovo}.

In our experiments, the database inside PUT MC contained $1530$ images and the $c_l$ was set to 40.
For those settings, the sequence matching time takes 150 ms for SGN3 and 105 ms for LS8-50.
This suggests that the system on SGN3 can process approx. 7 images per second which are slightly
less than what we recorded (8 fps). On LS8-50 device, the processing in real-time can be performed
for almost 10 frames per second, which leaves some processing time of CPU for other tasks in our case.
But the PUT MC database is relatively short when considering real-life applications and the
sequence matching time grows linearly with the number of images in the database. Therefore,
the system can be applied either in small buildings or it requires additional information
from an independent source in order to limit the search area to a single floor or a part of the building \cite{fabmapwifi}.
This might be an important factor when choosing between a typical place recognition
algorithm (e.g. FAB-MAP) and OpenABLE, as such a limitation was not observed for the OpenFABMAP.

\section{Nordland dataset}
One of the most challenging datasets for visual place recognition is the Nordland dataset \cite{seqslam}.
It was conceived for evaluation of outdoor place recognition under appearance changes induced by seasons
and consists of video sequences recorded during a train ride of approximately 3300 km.
The four sequences in Nordland correspond to four seasons -- spring, summer, fall and winter.
This dataset was used to evaluate the ABLE-M algorithm by its authors~\cite{able}.
The comparison window equal to 300 frames is assumed in \cite{able} and the processing
time is measured on a PC with i7 2.4 GHz CPU. 
We used the same settings to generate the
results presented in Tab.~\ref{tab:nordland}. The length of the testing sequence was to 1000 frames as in our previous experiments.

\begin{table}[htbp!]
\centering
\caption{Comparison of sequence matching time per frame for $c_l$ equal to 300 on Nordland for a typical i5 in a laptop (PC), Samsung Galaxy Note 3 (SGN3) and Lenovo S8-50 (LS8-50)}
\label{tab:nordland}
\begin{tabular}{c|ccc}
\multirow{2}{*}{Database size $n$} & \multirow{2}{*}{\begin{tabular}[c]{@{}c@{}}OpenABLE \\ PC\end{tabular}} & \multirow{2}{*}{\begin{tabular}[c]{@{}c@{}}OpenABLE \\ SGN3\end{tabular}} & \multirow{2}{*}{\begin{tabular}[c]{@{}c@{}}OpenABLE \\ LS8-50\end{tabular}} \\
                  &                                                                         &                                                                          &                                                                             \\ \hline
1000              & 0.169 s                                                                  & 0.627 s                                                                    & 0.366 s                                                                      \\
10 000            & 2.273 s                                                                  & 8.593 s                                                                   & 5.069 s                                                                      \\
100 000           & 24.476 s                                                                  & 85.480 s                                                                  & 52.103 s                                                                    
\end{tabular}
\end{table}

The authors of ABLE-M present the matching times when the training sequences are equal to
$1000$, $10000$ and $100000$ frames. For those sequences, the ABLE-M implementation in \cite{able}
takes 0.023 s, 0.25 s and 2.53 s per frame, respectively. Those times are much shorter comparing to the results
that we obtained on a laptop with i5 CPU, as we got, respectively, 0.169 s, 2.273 s and 24.476 s.
The difference in processing time might stem just from using a different computer, but we also do not
know how exactly the time was measured in \cite{able}. On mobile devices, Samsung Galaxy Note 3
and Lenovo S8-50, the obtained results are 2 to 4 times slower than those computed on the PC.

The results obtained on mobile devices demonstrate that OpenABLE cannot operate in real-time.
The solution proposed in~\cite{able} is to use multi-probe LSH (Local Sensitive Hashing)~\cite{lsh}
to reduce the matching time. The proposed version with LSH matching obtained 0.0093 s, 0.17 s and 0.42 s
for databases of 1000, 10 000, 100 000 images, respectively, but it provided only an approximate
solution that might return worse recognition results than the original algorithm, which was not verified in~\cite{able}.

\section{Improved FastABLE for Mobile Devices}
The ABLE-M algorithm presents an interesting idea exploiting sequences of images for visual place recognition.
Nevertheless, the OpenABLE implementation scales poorly with the number of frames stored in the database
and the length of the comparison window. An in-depth analysis of the ABLE-M algorithm enabled us to conclude
that there is a possibility to speed up the processing without any loss of recognition accuracy.
The improved algorithm has been implemented for PC/Linux and for Android mobile devices and is called FastABLE.

\subsection{Increasing the matching speed of OpenABLE}
\begin{figure}[htbp!]
 \includegraphics[width=\columnwidth]{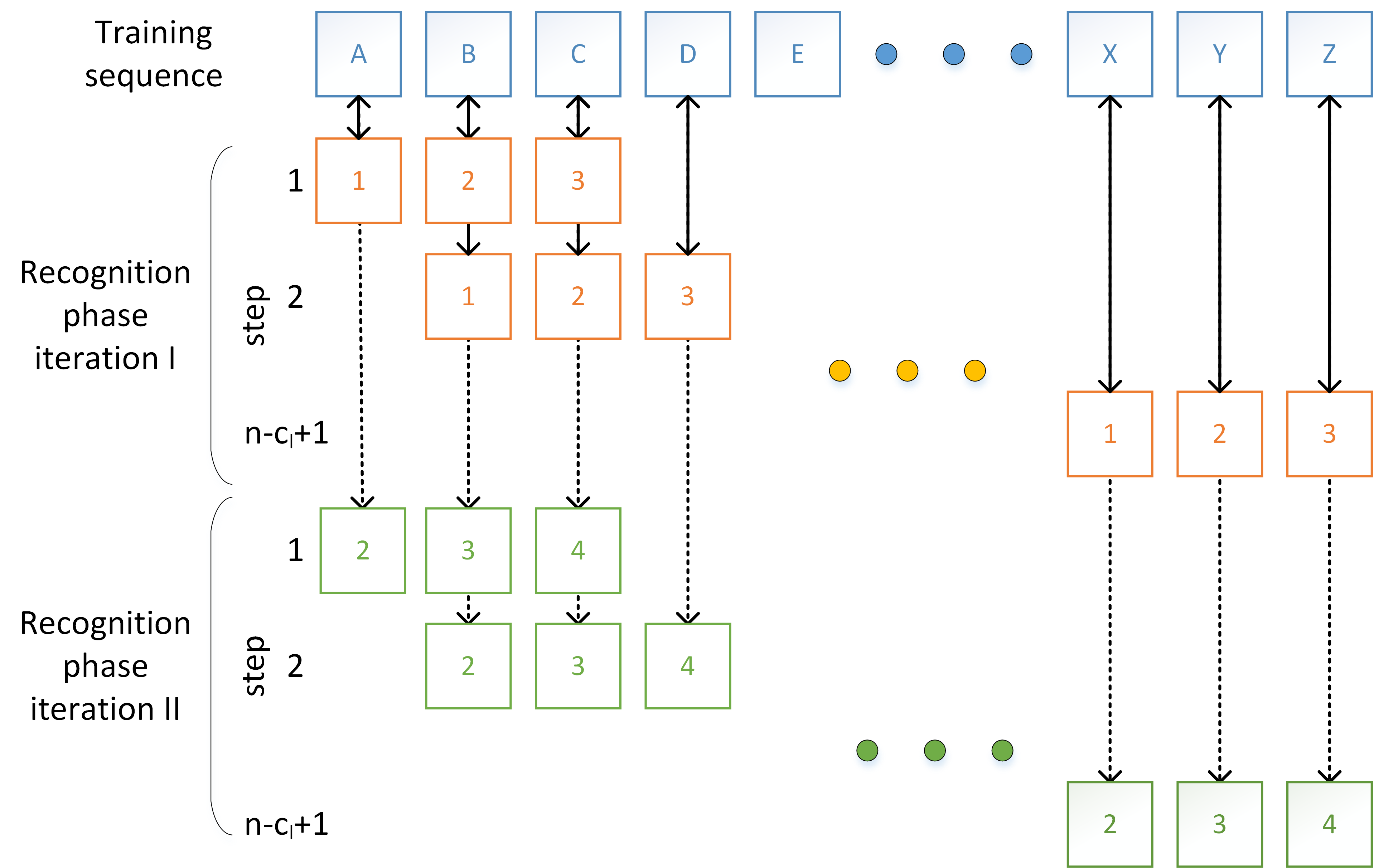}
 \caption{After capturing the image, one iteration with $n$ steps is performed to
         compare the sliding window of descriptors to the pre-computed descriptors from the database.
         In ABLE-M, each iteration is independent of the others.
         In FastABLE we utilize the previously computed values to speed-up the computations}
 \label{fig:computationScheme}
\end{figure}

The most important modification in the FastABLE implementation aims
at significantly increasing the speed of the matching phase and therefore increasing the related energy efficiency \cite{iciar}.
Although ABLE-M can employ the FLANN library \cite{flann} for fast nearest neighbor
search \cite{able}, this is an approximate technique, which improves processing
time mainly for very long sequences and is not implemented in OpenABLE. Thus,
we propose a purely algorithmic modification, that enables the use of our system
for real-time indoor localization on mobile devices. To introduce the idea,
we analyze the steps taken during processing in ABLE-M (Fig.~\ref{fig:computationScheme}).

In ABLE-M and the OpenABLE implementation, after a new image is captured,
the window of descriptors computed on the last $c_l$ images is moved along
the pre-recorded sequence and the corresponding Hamming distances are computed.
The distance computations between the same pairs of global descriptors in the
database sequence and the sliding window are repeated. This can be observed in
Fig.~\ref{fig:computationScheme}, as for example the distance between the
descriptors $B$ and $2$ is computed in iteration 1 step 1, and in iteration 2 step 2.

In contrary, in FastABLE we maximally re-use the information already computed.
In the first iteration, the corresponding distances are computed and saved in
an auxiliary array $prev$. In the next iterations, we compute the distance for
the first step the same way as in OpenABLE. However, in the following $n-1$ steps
it is possible to use the previously stored distance values. The distance $d$
for the iteration $j+1$ and the $(i+1)$-th step is computed as:
\begin{equation}
\begin{aligned}
    d_{i+1,j+1} = d_{i,j}+{\rm hamming}(train[i+c_l],test[j+c_l])\\ -{\rm hamming}(train[i],test[j]),
\end{aligned}
\end{equation}
where the sequence frames are indexed from 1. The idea is to compute the current distance as the result of
the previous iteration reduced by a distance for the first element in the previous comparison,
but increased by a distance for last element in the current sliding window. As we manipulate integer Hamming distances, no residual errors
accumulate due to the computations, and the results are numerically identical with those obtained from OpenABLE.
In Fig.~\ref{fig:computationScheme} the distance from the second step in second iteration, $d_{2,2}$, is
the distance between substring $\{B,C,D\}$ from pre-recorded sequence and substring $\{2,3,4\}$ from
the query sequence. Distance $d_{2,2}$ according to FastABLE is computed by taking the distance $d_{1,1}$
(distance between substring $\{A,B,C\}$ and substring$\{1,2,3\}$), subtracting the distance between $A$ and $1$
and increasing the result by the distance between $D$ and $4$. Each new distance updates the array $prev$,
so that in the next iteration it is possible to re-use the previous results.

The computational complexity of FastABLE is equal to $\mathcal{O}(n m)$,
as we perform computations for each new image in the recognition sequence ($m$),
and we slide the window $n$ times, but the comparison is performed in constant
time ($\mathcal{O}(1))$. Therefore, it is evident that the complexity of the
modified algorithm became independent of the length of the comparison window $c_l$,
comparing it to the $\mathcal{O}(n m c_l)$ complexity of OpenABLE. The computations
are done at an increased memory cost to store the distances computed in the previous iteration.

\subsection{Processing time analysis of FastABLE}
The proposed algorithm was implemented and is available on GitHub in a version
for PC\footnote{https://github.com/LRMPUT/FastABLE}, and in a version for Android devices\footnote{https://github.com/LRMPUT/FastABLE\_Android}.

\begin{figure}[htbp!]
 \centerline{\includegraphics[width=0.9\columnwidth]{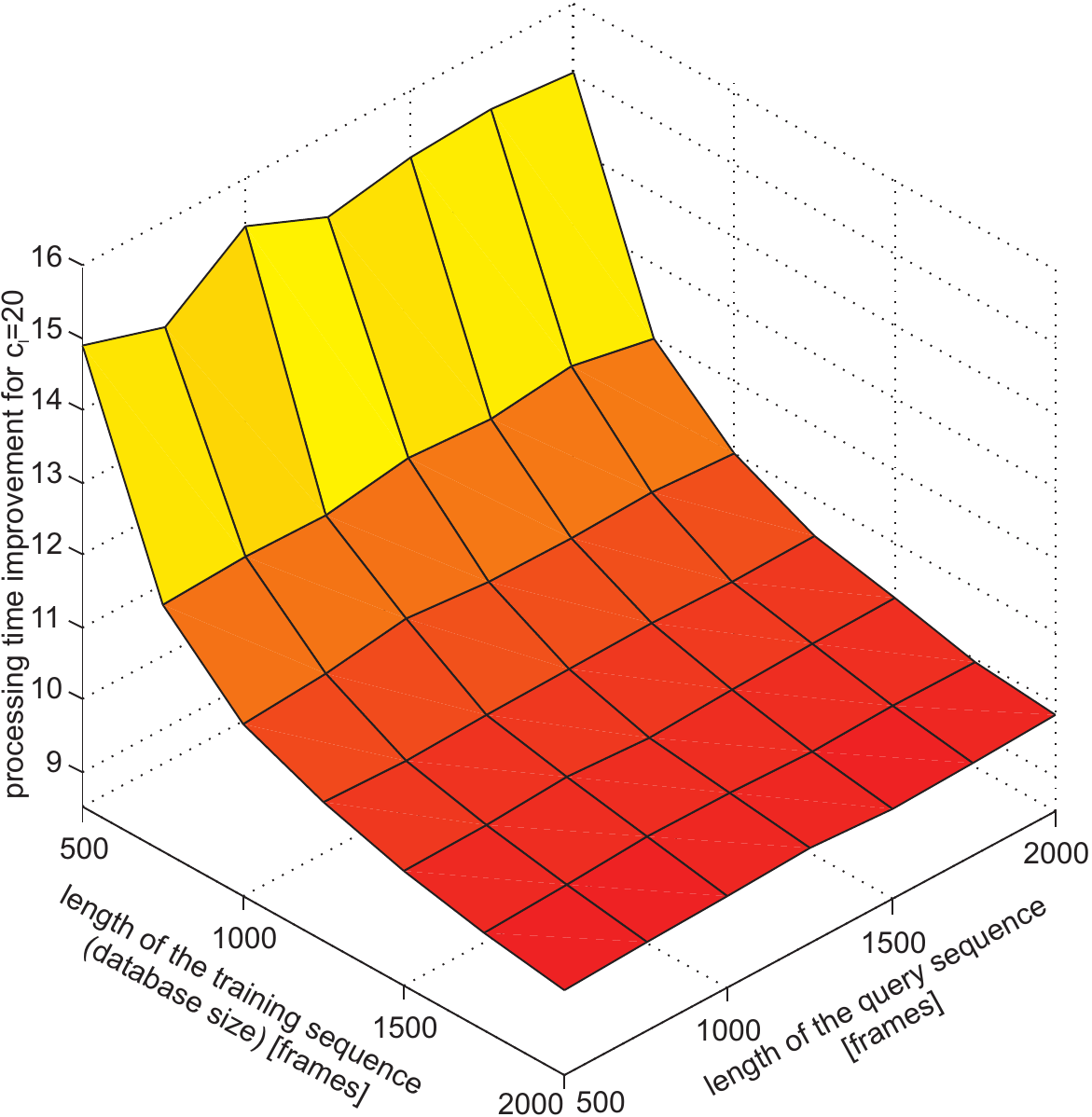}}
  \caption{Dependence of the speed improvement of FastABLE over OpenABLE
           from the length of the database and the legth of the query sequence on Samsung Galaxy Note 3 with $c_l$ equal to 20}
 \label{fig:size2speed}
\end{figure}

We started analysing the FastABLE performance from a test that verified how much the FastABLE algorithm execution time
depends on the length of the used image sequences: the database size, and the query sequence length, which, assuming a constant frame rate of the camera,
is proportional to the distance covered by the user. We performed an experiment for for an arbitrary chosen window length $c_l$=20,
and increasing lengths of the training database and the query sequence (Fig.~\ref{fig:size2speed}).
For real-life scenarios the window $c_l$ is usually much shorter than the length of the query sequence,
and therefore we present results for query sequences of length from 500 to 2000 frames.
From Fig.~\ref{fig:size2speed} it is evident that the speed improvement of FastABLE over OpenABLE is almost the same for
different lengths of the query sequence. Small differences are attributed to the non-deterministic execution time
of the multi-thread application. Considering these results, we assume the query sequence of 1000 frames for all further experiments.

Next, the FastABLE performance was evaluated by comparing the processing time
for different $c_l$ lengths, and different lengths of the database sequence.

\begin{figure}[htbp!]
 \includegraphics[width=\columnwidth]{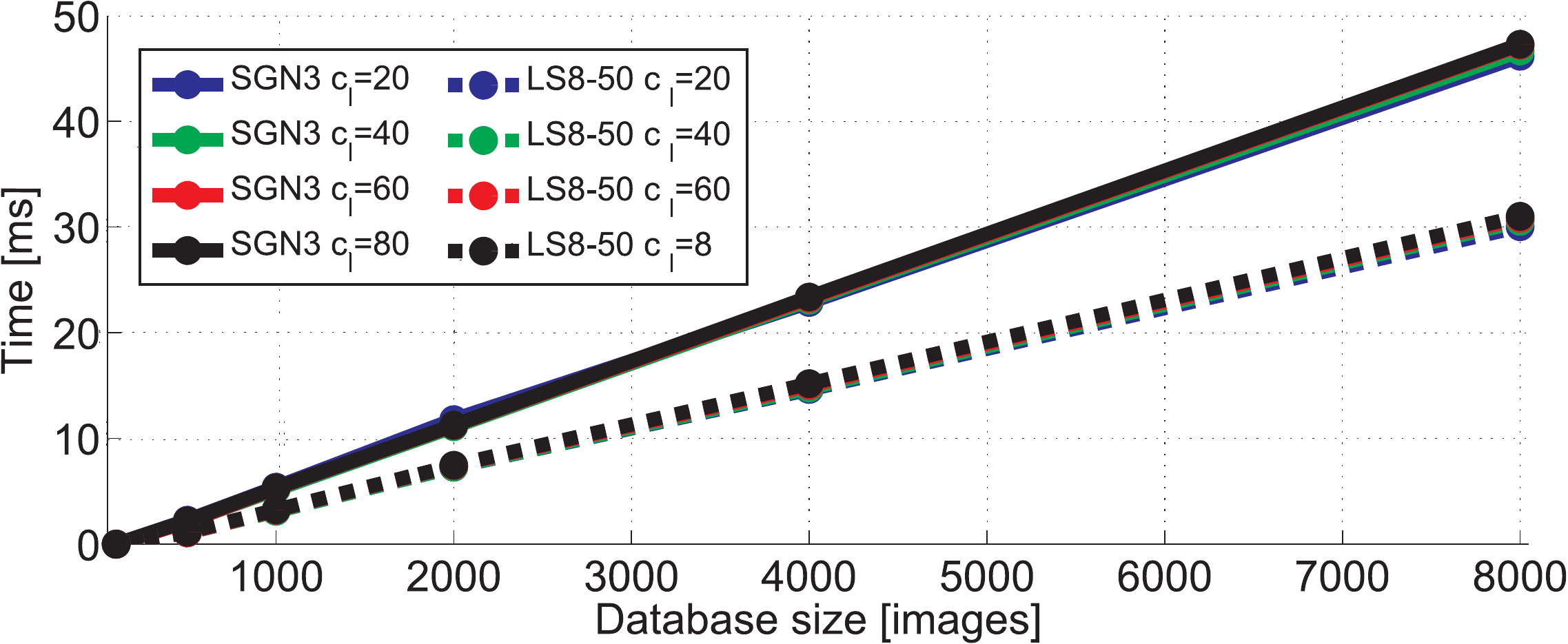}
  \caption{Time taken to match a new sequence of images to the pre-recorded database
           depending on the database size and the length of comparison window for FastABLE on Samsung Galaxy Note 3 (SGN3) and Lenovo S8-50L (LS8-50)}
 \label{fig:ableNewTime}
\end{figure}

Results presented in Fig.~\ref{fig:ableNewTime} prove that the size of the
sliding window $c_l$ does not have an influence on the processing time anymore.
Moreover, the obtained total processing times are much smaller than the times
required to process the same data by OpenABLE. The maximum processing
time took by FastABLE for a database of 4000 frames with $c_l$=80 is
below 24 ms, whereas OpenABLE needed 832 ms to accomplish the same task.
Table~\ref{tab:times} shows how many times faster is FastABLE with respect
to OpenABLE for several database sizes and $c_l$ values. Gains due to the
improved algorithm are especially evident for longer comparison windows.

\begin{table}[htbp!]
\centering
\caption{Processing time improvement over OpenABLE obtained with FastABLE on Samsung Galaxy Note 3 (SGN3) and Lenovo S8-50 (LS8-50)}
\label{tab:times}
\begin{tabular}{cc|ccccc}
                               &     & \multicolumn{5}{c}{database size $n$}                                                                                                                                         \\
                               & $c_l$ &  500     & 1000   & 2000  & 4000   & 8000   \\ \hline
\multirow{4}{*}{SGN3}          & 20    &  11.4 & 9.8 & 9.3  & 9.2 & 9.2 \\
                               & 40                                            & 22.3                     & 19.5                    & 18.6                    & 18.2                    & 18.2                    \\
                               & 60                                            & 32.5                     & 28.0                    & 27.5                    & 27.1                    & 26.9                    \\
                               & 80                                           & 40.5                     & 36.5                    & 36.2                    & 35.6                    & 35.6                    \\ \hline
\multirow{4}{*}{LS8-50} & 20                                            & 15.3                     & 11.0                    & 9.7                     & 9.7                     & 9.5                     \\
                               & 40                                         & 29.6                    & 21.7                    & 19.2                    & 19.2                    & 18.7                    \\
                               & 60                                            & 42.3                     & 31.2                    & 28.1                    & 28.1                    & 27.6                    \\
                               & 80                                            & 52.2                     & 40.2                    & 36.6                    & 37.1              & 36.5
\end{tabular}
\end{table}

\begin{figure}[htbp!]
 \includegraphics[width=\columnwidth]{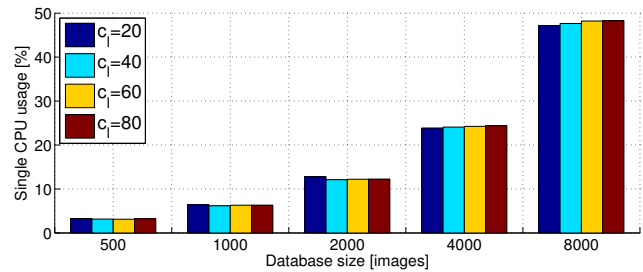}
  \caption{The usage of the single CPU depending on the database size and the length of
           comparison window for FastABLE on Samsung Galaxy Note 3 when images are processed at 10 Hz}
 \label{fig:faCPU}
\end{figure}

Similarly to the previous analysis, we can determine the maximal number of frames
that can be processed in real-time. As FastABLE can work in real-time on a smartphone,
we present those results with respect to the CPU usage, assuming that images from the
smartphone's camera are taken at 10 Hz, which simulates a realistic scenario for localization.
The results obtained for SGN3 are presented in Fig.~\ref{fig:faCPU}.

\begin{figure}[htbp!]
 \includegraphics[width=\columnwidth]{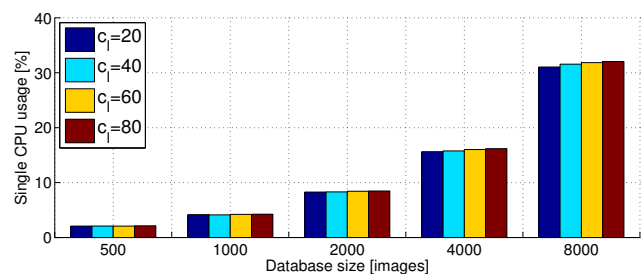}
  \caption{The usage of the single CPU depending on the database size and the length of comparison
           window for FastABLE on Lenovo S8-50 when images are processed at 10 Hz}
 \label{fig:faCPULenovo}
\end{figure}

The FastABLE on SGN3 can easily work in the background for small database sizes
or can process huge databases at the cost of high CPU usage. Similar results
are obtained for the LS8-50 device (Fig.~\ref{fig:faCPULenovo}). As we consider a
mobile device application, we recommend recording short databases and keeping the CPU usage low,
to avoid a negative influence on battery life.

\subsection{Verifying the gains of FastABLE on the Nordland dataset}
The possible gains of FastABLE were also evaluated on the long Nordland dataset.
We wanted to compare the FastABLE matching time with the OpenABLE implementation.
The experiments were conducted for a query sequence containing 1000 frames,
and are presented in Tab.~\ref{tab:nordland2}. In this experiment, FastABLE
proved to be at least 100 times faster than the OpenABLE implementation
regardless of the length of the database sequence.

\begin{table}[htbp!]
\centering
\caption{Single sequence matching times for $c_l$ equal to 300 on Nordland dataset}
\label{tab:nordland2}
\begin{tabular}{c|ccc}
Database size   & 1000   & 10 000 & 100 000 \\ \hline
OpenABLE PC     & 0.169 s & 2.273 s & 24.476 s  \\
FastABLE PC     & 0.0016 s & 0.019 s & 0.205 s  \\ \hline
OpenABLE SGN3   & 0.627 s & 8.593 s & 85.480 s \\
FastABLE SGN3   & 0.0056 s & 0.075 s & 0.751 s  \\ \hline
OpenABLE Lenovo & 0.366 s & 5.069 s & 52.103 s \\
FastABLE Lenovo & 0.0036 s & 0.043 s & 0.442 s
\end{tabular}
\end{table}

When considering the PC version, FastABLE is even faster than the reference
implementation of ABLE-M with the multi-probe LSH presented in \cite{able},
as for a database of 100000 images FastABLE performs a single matching in 205 ms,
while ABLE-M with LSH requires 420 ms, as stated in \cite{able}. Moreover,
FastABLE provides exact results, whereas the LSH version returns only an approximation.
It should be noted, that the FastABLE results were obtained on a slower PC
than the results presented in~\cite{able}. When it comes to mobile devices,
the FastABLE algorithm significantly speeds up the processing, but ap\-pear\-ance\--based
localization still cannot be accomplished in real-time for huge databases,
e.g. containing images from several multi-floor buildings. Therefore,
we recommend to use FastABLE and additionally divide a huge database
into smaller sequences utilizing prior information from other sources \cite{fabmapwifi}.

\subsection{Automatic tuning of FastABLE parameters}
The FastABLE parameters $t_r$ and $c_l$ can be set manually to maximize
the number of correct recognitions without false positives, but we propose to automatize this process.
The parameter $c_l$ should be large enough to ensure robustness to self-similarity of the environment,
but cannot exceed the length of the shortest sequence used as query. Due to
crossings inside the building, the pre-recorded video inside building consists of $p$ non-overlapping
sequences. Knowing the value of $c_l$ we match each of those sequences against the
remaining $p-1$ pre-recorded sequences to find the minimal Hamming distance between
the corresponding sub-sequences. As neither of the $p$ parts overlap, no recognitions
should be found, and $t_r$ is computed individually for the $a$-th sequence:
\begin{equation}
    t_r^a = \min_{i,j,b \neq a} {\rm hamming}(train_a[i,i+c_l-1],train_b[j,j+c_l-1]),
\end{equation}
where $t_r^a$ is the distance threshold found for $train_a$, which is part $a$
of the training sequence. This solution sets different thresholds for different
parts of the database, as different areas in the environment can differ significantly in their appearance.

\begin{figure}[thpb!]
\centering
 \includegraphics[width=\columnwidth]{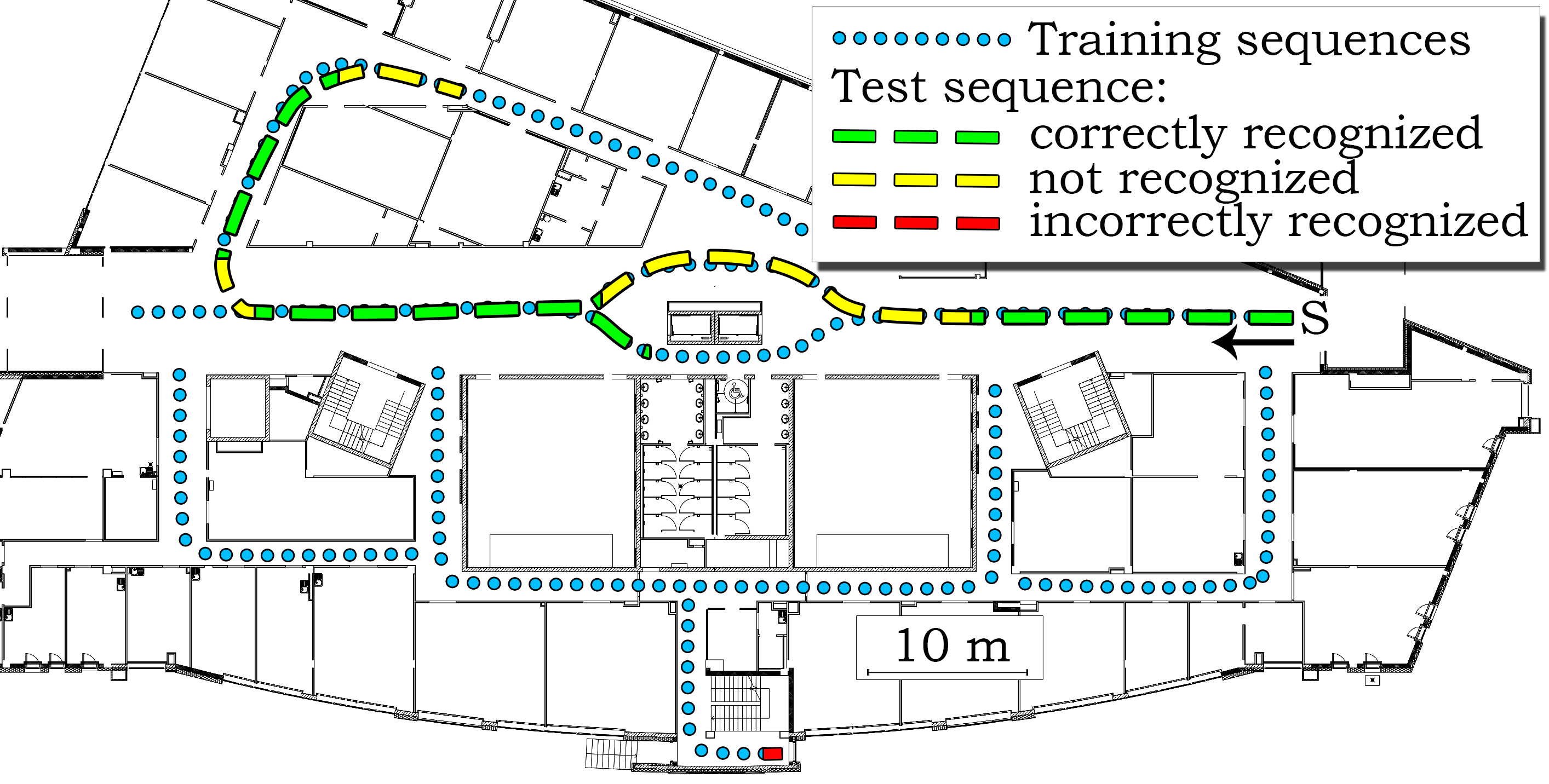}
 \caption{FastABLE experimental evaluation in the PUT MC building.
          The 14 pre-recorded sequences (blue dots) cover corridors in both directions.
          The query path (rectangles) consists of mostly correctly
          recognized locations (green) with some parts not recognized (yellow),
          and few incorrect (red)}
\label{fig:CM}
\end{figure}

\begin{figure}[thpb!]
\centering
 \includegraphics[width=\columnwidth]{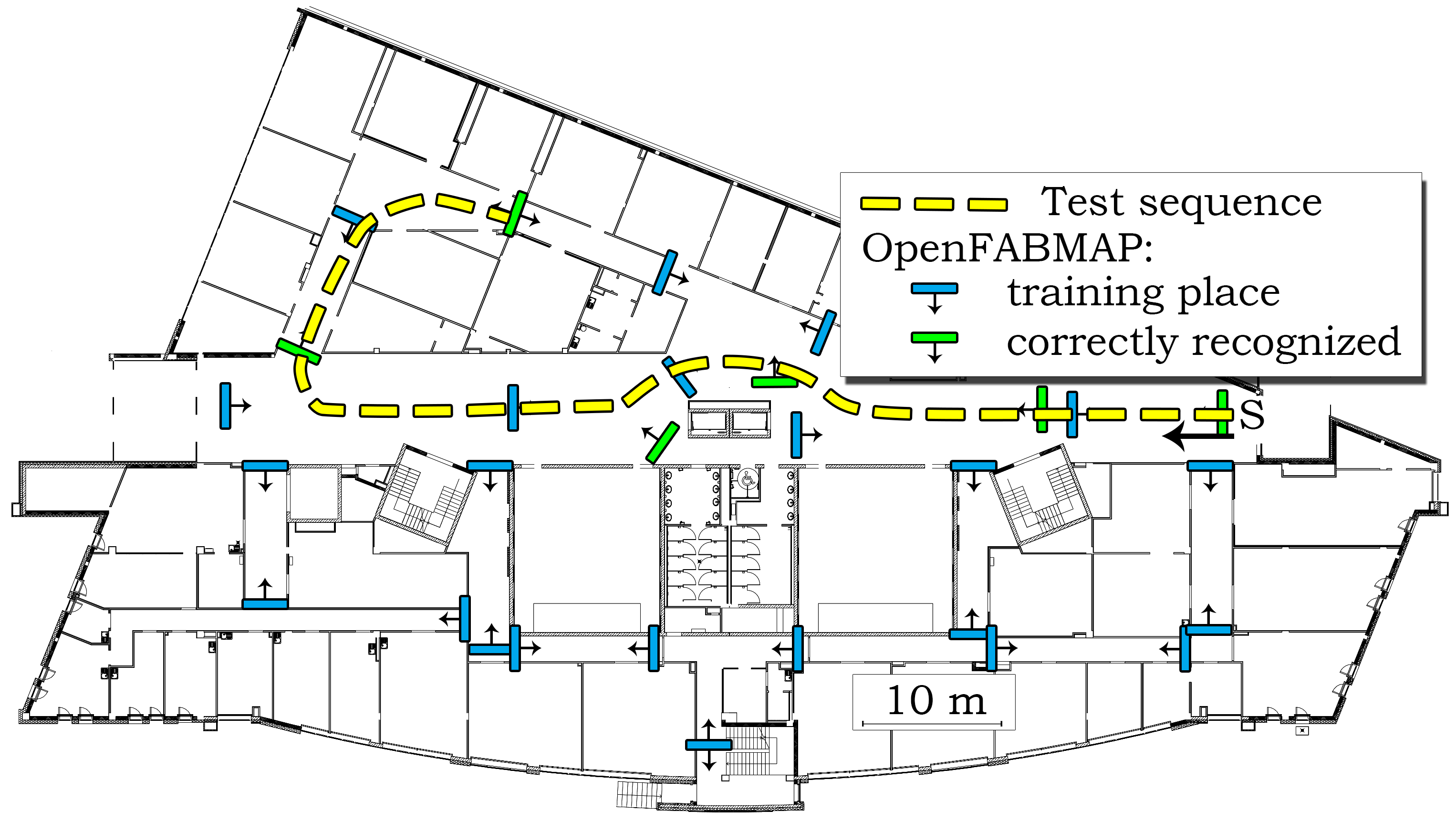}

 \caption{OpenFABMAP experimental evaluation in the PUT MC building.
          From 30 training recognition places (blue), OpenFABMAP correctly
          recognized 6 query locations (green) along the test path}
\label{fig:CM2}
\end{figure}

To verify the feasibility of FastABLE for indoor localization on mobile devices,
we performed experiments on the first floor of the Poznan University of Technology
Mechatronics Centre (PUT MC) building. The pre-recorded data consisted of 14 sequences
recorded on non-overlapping paths in both directions of motion. The database
amounted to 2127 images, while the query trajectory consisted of 320 images (Fig.~\ref{fig:CM}).
FastABLE reported 5625 correct place recognitions localizing the user on over a half of the trajectory, only with the camera images.
FastABLE returned 7 incorrect recognitions informing about localization in another part of the building that could be further removed by checking the consistency of localizations. The results were obtained with
automatically set parameters ($c_l$,$t_r$), and could be further improved by manual
tuning for the specific environment. 
The same query sequence was processed by the OpenFABMAP implementation of
the FAB-MAP algorithm with minor modifications (as presented in section \ref{fabmapa}).
OpenFABMAP correctly recognized user locations at 6 places along the path
from 30 training places (Fig.~\ref{fig:CM2}), but provided only sparse
location information that might not be suitable for some applications.

\section{Conclusions}
The article investigates the suitability of the ap\-pear\-ance\--based visual place
recognition methods for personal localization implemented on handheld mobile devices.
We focus on indoor environments, especially large buildings with multiple corridors
and many self-similar locations. We test two approaches: the ``golden standard''
place recognition algorithm FAB-MAP, and the more recent ABLE-M algorithm, which utilizes image sequences.
In the challenging environment, the open-source OpenFABMAP or OpenABLE implementations
do not provide reliable results. Therefore, we propose a simple modification that allows
OpenFABMAP to use a short sequence of images, which results in a working solution.
The OpenABLE implementation is also slightly modified, by clustering the results to ease
their interpretation in the context of personal navigation. From the experimental evaluation,
we conclude that both algorithms can be used for indoor visual place recognition
with a handheld camera, such as the one in a smartphone. The ABLE-M algorithm directly
using sequences of registered images proved to work reliably in the target environments,
working correctly when people were moving in front of the camera or were occluding
parts of the query images, and providing continuous localization over long trajectories.
These features made ABLE-M our preferred candidate to the ap\-pear\-ance\--based personal
localization solution.

However, another important aspect of a real-life application on mobile devices is
real-time processing and the related battery power consumption. In our experiments,
both algorithms were able to operate in real-time, but the processing time depended
on different factors for each of them. In FAB-MAP, the single image processing is
time-consuming, while the database matching is fast. In ABLE-M, the single image
processing is fast, while the database matching becomes time-consuming for longer
database sequences and longer comparison windows. This means, that ABLE-M is
potentially faster than FAB-MAP, but the processing time scales poorly with the
size of the environment and the length of user paths.

Therefore, in this paper, we propose the FastABLE solution, which is an implementation of
the modified ABLE-M algorithm that allows to speed-up the computations over 20 times
in typical indoor environments and to achieve low power consumption on the mobile device.
An open source version of FastABLE is publicly available allowing other researchers to
reproduce the presented results. The FastABLE algorithm allows to significantly increase
the size of the environment that the localization system can handle in real-time
or to increase the energy consumption efficiency for an environment of the same size.
The place recognition efficiency of FastABLE is the same as the baseline algorithm,
as the speed improvement is achieved by re-using efficiently the already computed information,
rather than by making approximate computations.

Our future research will focus on integrating FastABLE in the graph-based
multi-sensor personal localization scheme for mobile devices we have introduced in~\cite{mobicase}.


\bibliographystyle{spmpsci}       
\bibliography{wpc}                

\end{document}